\pdfoutput=1

\documentclass[11pt]{article}
\usepackage{authblk}
\usepackage[final]{EMNLP2022}

\usepackage{amsmath}
\usepackage{algpseudocode}
\usepackage{algorithm} 
\usepackage{algorithmicx}

\usepackage{times}
\usepackage{latexsym}
\usepackage[T1]{fontenc}
\usepackage[utf8]{inputenc}
\usepackage{microtype}
\usepackage{booktabs}
\usepackage{multirow}
\usepackage{svg}
\usepackage{hyperref}
\usepackage{pifont}
\usepackage{lipsum}
\usepackage{enumitem}
\usepackage{graphicx}
\usepackage{subcaption}

\algdef{SE}[DOWHILE]{Do}{doWhile}{\algorithmicdo}[1]{\algorithmicwhile\ #1}
\algnewcommand\algorithmicinput{\textbf{Input:}}
\algnewcommand\pseudoINPUT{\item[\algorithmicinput]}
\algnewcommand\algorithmicoutput{\textbf{Output:}}
\algnewcommand\pseudoOUTPUT{\item[\algorithmicoutput]}

\usepackage{linguex}

\usepackage{subcaption}

\usepackage{array}
\usepackage{graphics}

\setlist[itemize]{noitemsep, topsep=0pt}

\RequirePackage{etoolbox}
\RequirePackage{xcolor}
\definecolor{Gray}{gray}{0.9}
\definecolor{cb-black}      {RGB}{ 0,   0,   0}
\definecolor{cb-blue-green} {RGB}{ 0,  073,  073}
\definecolor{cb-green-sea}  {RGB}{ 0, 146, 146}
\definecolor{cb-rose}       {RGB}{255, 109, 182}
\definecolor{cb-salmon-pink}{RGB}{255, 182, 119}
\definecolor{cb-purple}     {RGB}{ 73,   0, 146}
\definecolor{cb-blue}       {RGB}{ 0, 109, 219}
\definecolor{cb-lilac}      {RGB}{182, 109, 255}
\definecolor{cb-blue-sky}   {RGB}{109, 182, 255}
\definecolor{cb-blue-light} {RGB}{182, 219, 255}
\definecolor{cb-burgundy}   {RGB}{146,   0,   0}
\definecolor{cb-brown}      {RGB}{146,  73,   0}
\definecolor{cb-clay}       {RGB}{219, 209,   0}
\definecolor{cb-green-lime} {RGB}{ 36, 255,  36}
\definecolor{cb-yellow}     {RGB}{255, 255, 109}

\newcommand{\xmark}{\ding{55}}
\newcommand\decreasespace{\vspace{-5pt}}

\title{Acceptability Judgements via Examining the Topology of Attention Maps}

\author{
    ~\textbf{Daniil Cherniavskii\textsuperscript{1,2}}\thanks{\ \ Equal contribution.},
    ~\textbf{Eduard Tulchinskii\textsuperscript{1}$^*$},
    ~\textbf{Vladislav Mikhailov\textsuperscript{3}$^*$}, \\
    ~\textbf{Irina Proskurina\textsuperscript{4}$^*$}, 
    ~\textbf{Laida Kushnareva\textsuperscript{5}},
    ~\textbf{Ekaterina Artemova\textsuperscript{5,7}}, \\
    ~\textbf{Serguei Barannikov\textsuperscript{1,6}}, 
    ~\textbf{Irina Piontkovskaya\textsuperscript{5}},
    ~\textbf{Dmitri Piontkovski\textsuperscript{4}},
    ~\textbf{Evgeny Burnaev\textsuperscript{1,2}} \\ 
    \textsuperscript{1}Skolkovo Institute of Science and Technology, 
    \textsuperscript{2}AIRI,
    \textsuperscript{3}SberDevices, \\
    \textsuperscript{4}HSE University,
    \textsuperscript{5}Huawei Noah’s Ark lab, \textsuperscript{6}CNRS, IMJ \\ 
    \textsuperscript{7}Center for Information and Language Processing (CIS), LMU Munich, Germany \\
    
    \small{
    \textbf{Correspondence:} \href{mailto:Eduard.Tulchinskiy@skoltech.ru}{Eduard.Tulchinskiy@skoltech.ru}}
}

\begin{document}

\maketitle

\begin{abstract}
The role of the attention mechanism in encoding linguistic knowledge has received special interest in NLP. However, the attention heads' ability to judge the grammatical acceptability of a sentence has been underexplored. This paper approaches the paradigm of acceptability judgments with topological data analysis (TDA), showing that the topological properties of the attention graph can be efficiently exploited for two standard practices in linguistics: binary judgments and linguistic minimal pairs. Topological features enhance the BERT-based acceptability classifier scores by up to $0.24$ Matthew's correlation coefficient score on \textsc{CoLA} in three languages (English, Italian, and Swedish). By revealing the topological discrepancy between attention graphs of minimal pairs, we achieve the human-level performance on the \textsc{BLiMP} benchmark, outperforming nine statistical and Transformer LM baselines. At the same time, TDA provides the foundation for analyzing the linguistic functions of attention heads and interpreting the correspondence between the graph features and grammatical phenomena. We publicly release the code and other materials used in the experiments\footnote{\href{https://github.com/danchern97/tda4la}{\texttt{github.com/danchern97/tda4la}}}.
\end{abstract}

\section{Introduction}
Linguistic competence of neural language models (LMs) has emerged as one of the core sub-fields in NLP. The research paradigms explore whether Transformer LMs~\cite{vaswani2017attention} induce linguistic generalizations from raw pre-training corpora~\cite{warstadt-etal-2020-learning,zhang-etal-2021-need}, what properties are learned during task-specific fine-tuning~\cite{miaschi-etal-2020-linguistic,merchant-etal-2020-happens}, and how the experimental results are connected to grammar and language acquisition theories~\cite{pater2019generative,manning2020emergent}. %

One of these paradigms is centered around acceptability judgments, which have formed an empirical foundation in generative linguistics over the last six decades~\cite{chomsky1965aspects,schutze1996empirical,sep-linguistics}. Acceptability of linguistic stimuli is traditionally 
investigated in the form of a forced choice between binary categories or minimal pairs~\cite{sprouse2018acceptability}, which are widely adopted for acceptability classification~\cite{linzen2016assessing,warstadt-etal-2019-neural} and probabilistic LM scoring~\cite{lau2017grammaticality}.

A scope of approaches has been proposed to interpret the roles of hundreds of attention heads in encoding linguistic properties~\cite{htut2019attention,wu-etal-2020-perturbed} and identify how the most influential ones benefit the downstream performance~\cite{voita-etal-2019-analyzing,jo-myaeng-2020-roles}. Prior work has demonstrated that heads induce grammar formalisms and structural knowledge~\cite{zhou-zhao-2019-head,lin-etal-2019-open,luo-2021-attention}, and linguistic features motivate attention patterns~\cite{kovaleva-etal-2019-revealing,clark-etal-2019-bert}. Recent studies also show that certain heads can have multiple functional roles~\cite{pande2021heads} and even perform syntactic functions for typologically distant languages~\cite{ravishankar-etal-2021-attention}. 

Our paper presents one of the first attempts to analyze attention heads in the context of linguistic acceptability (LA) using topological data analysis (TDA\footnote{We also refer the reader to works on computational topology and its applications~\cite{Barannikov94,zomorodian2001computing,books/daglib/0025666,carlsson_vejdemo-johansson_2021}.};~\citealp{chazal2017introduction}). TDA allows for exploiting complex structures underlying textual data and investigating graph representations of Transformer's attention maps. We show that topological features are sensitive to well-established LA contrasts, and the grammatical phenomena can be encoded with the %
topological
properties of the attention map.

The main contributions are the following:~\emph{(i)}~ We adapt TDA methods to two standard approaches to LA judgments: acceptability classification and scoring minimal pairs (\S\ref{sec:methodology}).~\emph{(ii)}~We conduct acceptability classification experiments in three Indo-European languages (English, Italian, and Swedish) and outperform the established baselines (\S\ref{sec:acceptability_classification}).~\emph{(iii)}~We introduce two scoring functions, which reach the human-level performance in discriminating between minimal pairs in English and surpass nine statistical and Transformer LM baselines (\S\ref{sec:minimal_pairs}).~\emph{(iv)}~The linguistic analysis of the feature space proves that TDA can serve as a complementary approach to interpreting the attention mechanism and identifying heads with linguistic functions (\S\ref{subsection:cola_results},\S\ref{subsection:blimp_results},\S\ref{sec:discussion}). 

\section{Related Work}
\subsection{Linguistic Acceptability} 
\paragraph{Acceptability Classification.}
Early works approach acceptability classification with classic ML methods, hand-crafted feature templates, and probabilistic syntax parsers~\cite{cherry-quirk-2008-discriminative,wagner2009judging,post-2011-judging}. Another line employs statistical LMs~\cite{heilman-etal-2014-predicting}, including threshold-based classification with LM scoring functions~\cite{clark-etal-2013-statistical}. The ability of RNN-based models~\cite{elman1990finding,hochreiter1997long} to capture long-distance regularities has stimulated investigation of their grammatical sensitivity~\cite{linzen2016assessing}. With the release of the Corpus of Linguistic Acceptability (\textsc{CoLA};~\citealp{warstadt-etal-2019-neural}) and advances in language modeling, the focus has shifted towards Transformer LMs~\cite{yin-etal-2020-robustness}, establishing LA as a proxy for natural language understanding (NLU) abilities~\cite{wang-etal-2018-glue} and linguistic competence of LMs~\cite{warstadt2019linguistic}. %

\paragraph{Linguistic Minimal Pairs.} A forced choice between minimal pairs is a complementary approach to LA, which evaluates preferences between pairs of sentences that contrast an isolated grammatical phenomenon~\cite{schutze1996empirical}. The idea of discriminating between minimal contrastive pairs has been widely applied to scoring generated hypotheses in downstream tasks~\cite{pauls-klein-2012-large,salazar-etal-2020-masked}, measuring social biases~\cite{nangia-etal-2020-crows}, analyzing machine translation models~\cite{burlot-yvon-2017-evaluating,sennrich-2017-grammatical}, and linguistic profiling of LMs in multiple languages~\cite{marvin-linzen-2018-targeted,mueller-etal-2020-cross}.

\subsection{Topological Data Analysis in NLP} TDA has found several applications in NLP. One of them is word sense induction by clustering word graphs and detecting their connected components. The graphs can be built from word dictionaries~\cite{levary2012loops}, association networks~\cite{dubuisson-etal-2013-topology}, and word vector representations~\cite{jakubowski-etal-2020-topology}. Another direction involves building classifiers upon geometric structural properties for movie genre detection~\cite{doshi2018movie}, textual entailment~\cite{savle-etal-2019-topological}, and document classification~\cite{das2021persistence,werenski2022measure}. Recent works have mainly focused on the topology of LMs' internal representations. \citet{kushnareva-etal-2021-artificial} represent attention maps with TDA features to approach artificial text detection. \citet{colombo-etal-2021-automatic} introduce \textsc{BaryScore}, an automatic evaluation metric for text generation that relies on Wasserstein distance and barycenters. To the best of our knowledge, TDA methods have not yet been applied to LA.

\section{Methodology}
\label{sec:methodology}

\begin{figure*}[th!]
\centering
\begin{subfigure}{0.49\textwidth}
  \centering
  \caption{Attention map (left); Barcode (right); [L: $1$; H: $11$]. }
  \includegraphics[width=0.75\textwidth]{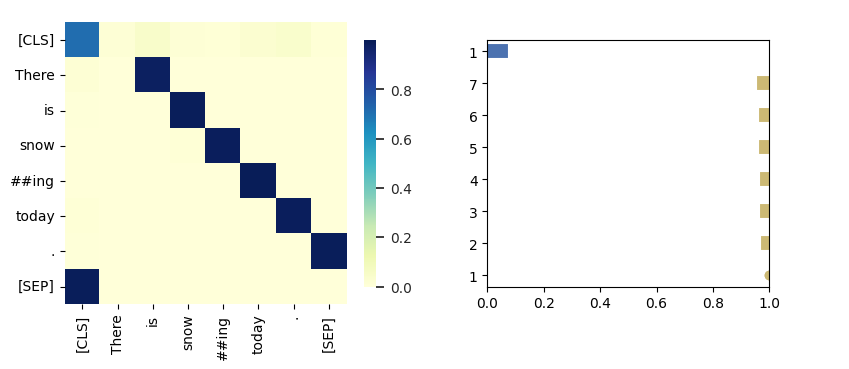}
  \label{fig:map_and_barcode_a}
\end{subfigure}
\hfill
\begin{subfigure}{0.49\textwidth}
  \centering
  \caption{Attention map (left); Barcode (right); [L: $9$; H: $9$].}
  \includegraphics[width=0.75\textwidth]{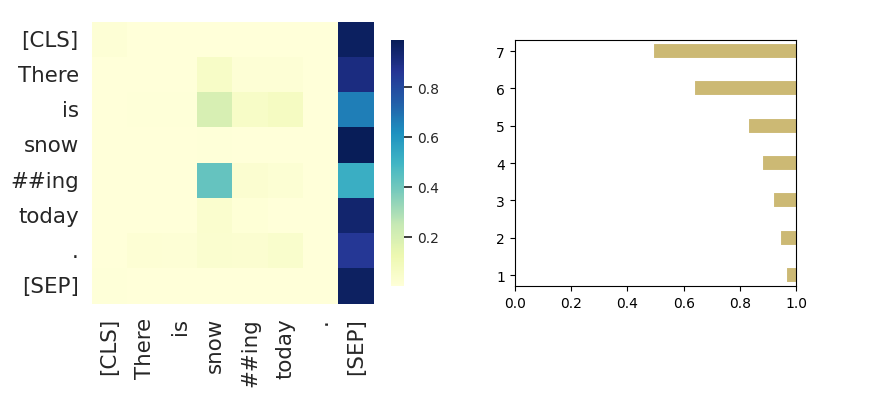}
  \label{fig:map_and_barcode_b}
\end{subfigure}
\begin{subfigure}{.95\textwidth}
  \centering
  \caption{Attention graph filtration: [L: $1$; H: $11$].}
  \includegraphics[width=0.95\textwidth]{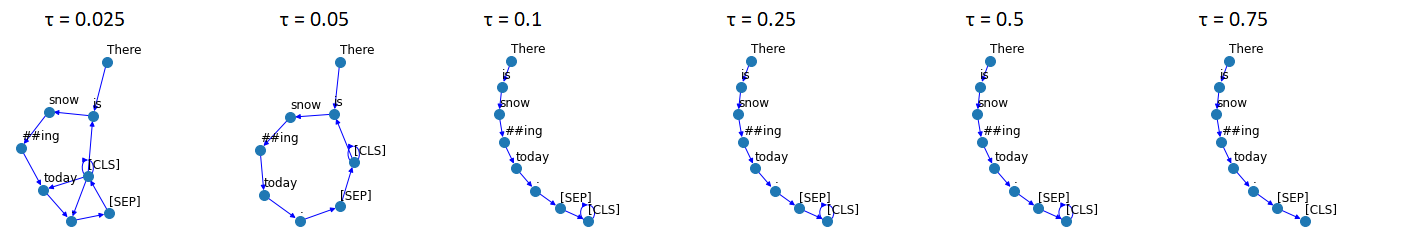}
  \label{fig:graphs_a}
\end{subfigure}
\begin{subfigure}{.95\textwidth}
  \centering
  \caption{Attention graph filtration: [L: $9$; H: $9$].}
  \includegraphics[width=0.95\textwidth]{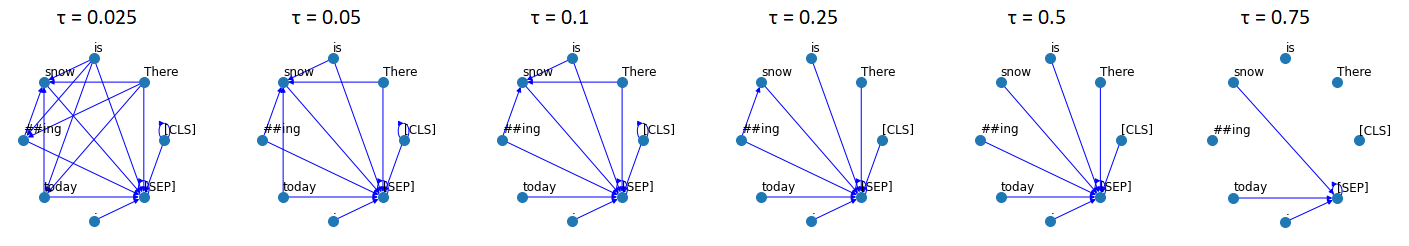}
  \label{fig:graphs_b}
\end{subfigure}
\caption{
An example of attention maps, barcodes, and filtration procedure for the sentence \emph{``There is snowing today.''}. Model=En-BERT-base~\cite{devlin-etal-2019-bert}. Heads=[Layer $1$; Head $11$] and [Layer $9$; Head $9$].
}
\label{fig:graph_examples}
\end{figure*}

\subsection{Attention Graph}
We treat Transformer's attention matrix $A^{attn}$ as a weighted graph $G$, where the vertices represent tokens, and the edges connect pairs of tokens with mutual attention weights. This representation can be used to build a family of attention graphs called \emph{filtration}, i.e., an ordered set of graphs $G^{\tau_i}$ filtered by increasing attention weight thresholds ${\tau_i}$. Filtering edges lower than the given threshold affects the graph structure and its core features, e.g., the number of edges, connected components, or cycles. TDA techniques allow tracking these changes, identifying the moments of when the features appear (i.e., their ``\emph{birth}'') or disappear (i.e., their \emph{``death''}), and associating a lifetime to them. The latter is encoded as a set of intervals called a \emph{``barcode''}, where each interval (\emph{``bar''}) lasts from the feature's ``birth'' to its ``death''. The barcode characterizes the persistent features of attention graphs and describes their stability.

\paragraph{Example.} Let us illustrate the process of computing the attention graph filtration and barcodes given an Example~\ref{eg:tda}. 
\decreasespace{}
\ex.  There is snowing today.
\label{eg:tda}  
\decreasespace{}

First, we compute attention maps for each Transformer head as shown in~\autoref{fig:map_and_barcode_a}-\ref{fig:map_and_barcode_b} (left). These two heads follow different attention patterns~\cite{clark-etal-2019-bert}: attention to the next token (Figure~\ref{fig:map_and_barcode_a}) and to the \texttt{[SEP]} token (\autoref{fig:map_and_barcode_b}). Next, we represent the map as a weighted graph, and conduct the filtration procedure for a fixed set of attention weight thresholds. The edges lower than each given threshold are discarded, which results in a set of six attention graphs with their maximum spanning trees (MSTs) becoming a chain (\autoref{fig:graphs_a};~$\tau$=$0.1$), and a star (\autoref{fig:graphs_b};~$\tau$=$0.5$). The families of attention graphs are used to compute persistent features (\S\ref{persistent_features}).

\autoref{fig:map_and_barcode_a}-\ref{fig:map_and_barcode_b} (right) depict barcodes for each family of graphs. The bars are sorted by length. The number of bars equals $|T|-1$, where $|T|$ is the number of tokens in the input sentence. The bars in yellow correspond to the $0$-dimensional features acquired from the edges of the MST. The bars in blue refer to $1$-dimensional features, which stand for non-trivial simple cycles. Such cycle appears in the first family (\autoref{fig:graphs_a};~$\tau$=$0.05$), which is shown as a blue bar in~\autoref{fig:map_and_barcode_a}. By contrast, 
there are no cycles in the second family (\autoref{fig:graphs_b}) and on the corresponding barcode.

\begin{figure*}[th!]
    \centering
    \includegraphics[width=\linewidth]{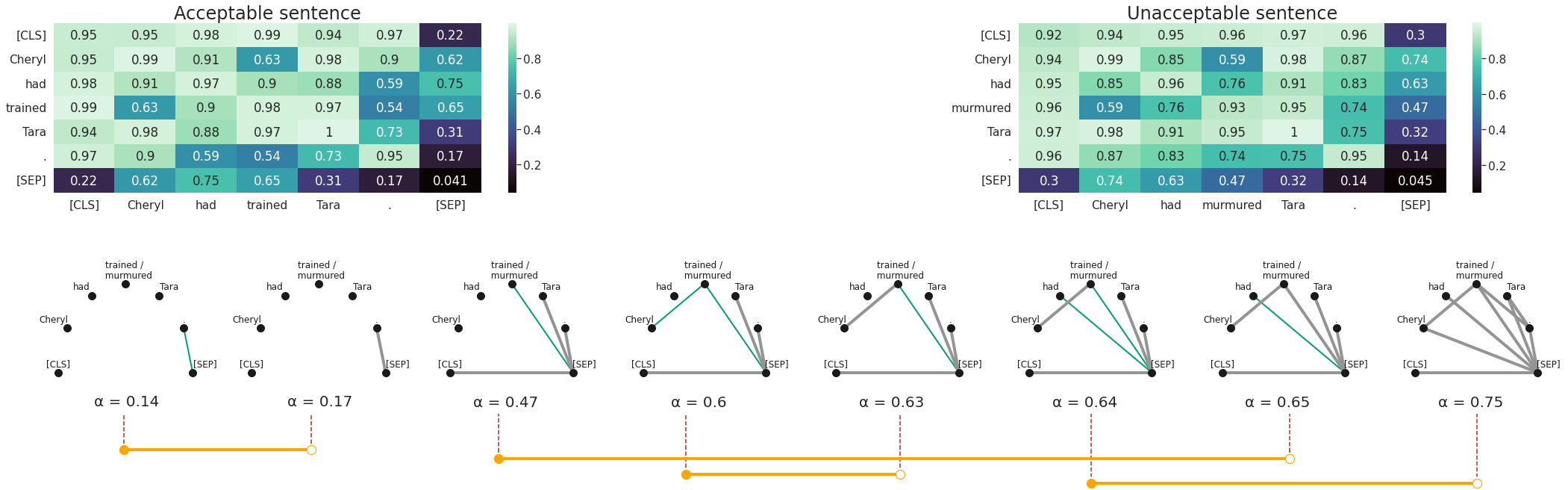}
    \caption{A graphical representation of RTD-barcodes. In the top row given $A^\prime$ matrices derived from attention maps for acceptable and unacceptable sentences. Edges present in both graphs $G_a^{\alpha_i}$ and $G_b^{\alpha_i}$ at a given threshold  ${\alpha_i}$ are colored in grey. Edges present only in graph $G_b^{\alpha_i}$ are colored in green.}
    \label{fig:rtd_clarification}
\end{figure*}

\subsection{Persistent Features of Attention Graphs}
\label{persistent_features}
We follow~\citet{kushnareva-etal-2021-artificial} to design three groups of persistent features of the attention graph: (i)~\textbf{topological features}, (ii)~\textbf{features derived from barcodes}, and (iii)~\textbf{features based on distance to attention patterns}.
The features are computed on attention maps produced by a Transformer LM.%

\paragraph{Topological Features.} Topological features include the first two Betti numbers of the undirected graph $\beta_0$ and $\beta_1$ and standard properties of the directed graph, such as the number of strongly connected components, edges, and cycles. The features are calculated on pre-defined thresholds over undirected and directed attention graphs from each head separately and further concatenated.

\paragraph{Features Derived from Barcodes.} Barcode is the representation of the graph's persistent homology~\cite{barannikov2021canonical}. We use the Ripser++ toolkit~\cite{zhang2020gpu} to compute $0$/$1$-dimensional barcodes for $A^{attn}$. Since Ripser++ leverages upon distance matrices, we transform $A^{attn}$ as $A'=1-\max{(A^{attn}, A^{attn~T})}$.

Next, we compute descriptive characteristics of each barcode, such as the sum/average/variance of lengths of bars, the number of bars with the time of birth/death greater/lower than a threshold, and the entropy of the barcodes. The sum of lengths of bars ($H_0S$)  corresponds to the $0$-dimensional barcode and represents the sum of edge weights  in the $A'$'s minimum spanning tree.
The average length of bars ($H_0M$) corresponds to the mean edge weight in this tree, i.e. $H_0M=1 - $(the mean edge weight of the maximum spanning tree in $A^{attn}$ ).

\paragraph{Features Based on Distance to Patterns.} The shape of attention graphs can be divided into several patterns: attention to the previous/current/next token, attention to the \texttt{[SEP]}/\texttt{[CLS]} token, and attention to punctuation marks~\cite{clark-etal-2019-bert}. We formalize attention patterns by binary matrices and calculate distances to them as follows. We take the Frobenius norm of the difference between the matrices normalized by the sum of their norms. The distances to patterns are used as a feature vector.

\paragraph{Notations.} We summarize the notations used throughout the paper: %
\begin{itemize}%
    \item $H_i$: $i$-th Homology Group 
    \item $\beta_i$: Betti number, dimension of $H_i$
    \item $H_0S$: The sum of lengths of bars
    \item $H_0M$: The average of lengths of bars
    \item $PCA$: Principal Component Analysis
    \item $PC^{\{i\}}$: Subset $\{i\}$ of principal components
    \item $MST$: Maximum Spanning Tree
    \item RTD: Representation Topology Divergence
\end{itemize}

\subsection{Representation Topology Divergence}
\label{rtd_method}
Representation Topology Divergence (RTD;~\citealp{barannikov2021representation}) measures topological dissimilarity between a pair of weighted graphs with one-to-one vertex correspondence. \autoref{fig:rtd_clarification} outlines computation of RTD for a sentence pair in Example~\ref{eg:rtd_pair}.

\decreasespace
\ex. \label{eg:rtd_pair} 
\a. Cheryl had trained Tara.
\b. * Cheryl had murmured Tara.
\decreasespace

First, we compute attention maps for the input sentences $S_a$ and $S_b$ with a Transformer LM, and represent them as the weighted graphs $G_a$ and $G_b$. Next, we establish a one-to-one match between the vertices, and sort the filtrations $G_a^{\alpha_i}$ and $G_b^{\alpha_i}$ with $\alpha$=$1-\tau$  in the ascending order. We then track the hierarchical formation of connected components in the graph $G_a^{\alpha_i} \cap G_b^{\alpha_i}$ while increasing $\alpha_i$. The RTD$(G_a,~G_b)$ feature appears at threshold $\alpha_{i}$ if an edge with the weight $\alpha_i$ in the graph $G_b$ joins two different connected components of the graph $G_a^{\alpha_{i}}\cap G_b^{\alpha_{i}}$. This feature disappears at the threshold $\alpha_j$ if the two $G_a^{\alpha_{i}}\cap G_b^{\alpha_{i}}$ connected components become joined in the graph $G_a^{\alpha_{j}}$. 

\paragraph{Example.} We can identify the ``birth'' of the RTD feature at $\alpha$=$0.47$, when an edge appears in $G_b^{\alpha=0.47}$ between the connected component ``\texttt{trained/murmured}'' and the connected component with four vertices, namely ``\texttt{[SEP]}'', ``\texttt{[CLS]}'', ``\texttt{.}'', and ``\texttt{Tara}'' (\autoref{fig:rtd_clarification}; the appearing edge is colored in green). We observe its ``death'', when the edge becomes present in both attention graphs at $\alpha$=$0.65$ (the corresponding edge changes its color to grey in the graph $G_a^{\alpha=0.65}\cap G_b^{\alpha=0.65}$). When comparing the graphs in this manner, we can associate a lifetime to the feature by computing the difference between the moments of its ``death'' (e.g., $\alpha_j$=$0.65$) and ``birth'' (e.g., $\alpha_i$=$0.47$). The lifetimes are illustrated as the orange bars $[\alpha_i,\alpha_j]$ in~\autoref{fig:rtd_clarification}. The resulting value of RTD$(G_a,~G_b)$ is the sum of lifetimes $\alpha_j-\alpha_i$ over all such features. A formal description of RTD is provided in~\autoref{sec:appendix_rtd}.

\section{Acceptability Classification}
\label{sec:acceptability_classification}
\subsection{Data}
We use three LA classification benchmarks in English~(\textsc{CoLA};~\citealp{warstadt-etal-2019-neural}), Italian~(\textsc{ItaCoLA};~\citealp{DBLP:journals/corr/abs-2109-12053}) and~Swedish (\textsc{DaLAJ};~\citealp{volodina-etal-2021-dalaj}). \textsc{CoLA} and \textsc{ItaCoLA} contain sentences from linguistic textbooks and cover morphological, syntactic, and semantic phenomena. The target labels are the original authors' acceptability judgments. \textsc{DaLAJ} includes L2-written sentences with morphological violations or incorrect word choices. The benchmark statistics are described in~\autoref{table:data_stats} (see~\autoref{sec:appendix_cola}). We provide examples of acceptable and unacceptable sentences in English~\autoref{eg:cola}, Italian~\autoref{eg:itacola}, and Swedish~\autoref{eg:dalaj} from the original papers.

\ex. \label{eg:cola} 
\a. What did Betsy paint a picture of?
\b. *Maryann should leaving.
\decreasespace

\decreasespace
\ex. \label{eg:itacola} 
\a. Ho voglia di salutare Maria. \\
``I want to greet Maria.''
\b. *Questa donna mi hanno colpito. \\
``This woman have impressed me.''
\decreasespace

\decreasespace
\ex. \label{eg:dalaj} 
\a. Jag kände mig jättekonstig. \\
``I felt very strange.''
\b. *Alla blir busiga med sociala medier. \\
``Everyone is busy with social media.''

\subsection{Models}
\label{la_models}
We run the experiments on the following Transformer LMs: En-BERT-base~\cite{devlin-etal-2019-bert}, It-BERT-base~\cite{stefan_schweter_2020_4263142}, Sw-BERT-base~\cite{swedish-bert}, and XLM-R-base~\cite{conneau-etal-2020-unsupervised}. Each LM has two instances: \emph{frozen} (a pre-trained model with frozen weights), and \emph{fine-tuned} (a model fine-tuned for LA classification in the corresponding language).

\paragraph{Baselines.}
We use the \emph{fine-tuned} LMs, and a linear layer trained over the pooler output from the \emph{frozen} LMs as baselines. 

\paragraph{Our Models.} We train Logistic Regression classifiers over the persistent features 
computed with each model instance: \emph{(i)}~the average length of bars ($H_0M$); \emph{(ii)}~concatenation of all topological features referred to as \textit{TDA} (\S\ref{persistent_features}). Following~\citeauthor{warstadt-etal-2019-neural}, we evaluate the performance with the accuracy score (Acc.) and Matthew's Correlation Coefficient (MCC;~\citealp{Matthews1975ComparisonOT}). The fine-tuning details are provided in~\autoref{sec:appendix_training}.

\subsection{Results} \label{subsection:cola_results}
\begin{table}[t!]
\scriptsize
\centering
\newcommand{\hsp}{\hspace{4pt}}
\setlength{\tabcolsep}{2pt}
\begin{tabular}{@{}lcccc@{}} 
\toprule
\multirow{3}{*}{\textbf{Model}} & \multicolumn{2}{c}{\textbf{Frozen LMs}} & \multicolumn{2}{c}{\textbf{Fine-tuned LMs}} \\[0.4ex]

& \textbf{IDD / Dev} & \textbf{OODD / Test} & \textbf{IDD / Dev} & \textbf{OODD / Test} \\ [0.4ex]

& \textbf{Acc.} \hsp \textbf{MCC} & \textbf{Acc.} \hsp \textbf{MCC} & \textbf{Acc.} \hsp \textbf{MCC} & \textbf{Acc.} \hsp \textbf{MCC} \\

\midrule

\multicolumn{5}{c}{\textbf{CoLA}} \\ 
\midrule
En-BERT & 69.6 \hsp 0.037 & 69.0 \hsp 0.082 & 83.1 \hsp 0.580 & 81.0 \hsp 0.536 \\
En-BERT + $H0M$ & \underline{75.0} \hsp \underline{0.338} & \underline{75.2} \hsp \underline{0.372} & 85.2 \hsp 0.635 & \underline{81.2} \hsp \underline{0.542} \\
En-BERT + \textit{TDA}  & \textbf{77.2} \hsp \textbf{0.420} & \textbf{76.7} \hsp \textbf{0.420} & \textbf{88.6} \hsp \textbf{0.725} & \textbf{82.1} \hsp \textbf{0.565} \\ 

XLM-R & 68.9 \hsp 0.041 & 68.6 \hsp 0.072 & 80.8 \hsp 0.517 & 79.3 \hsp 0.489 \\ 
XLM-R + $H0M$  & 71.3 \hsp 0.209 & 69.8 \hsp 0.187 & 81.2 \hsp 0.532 & 77.7 \hsp 0.445 \\
XLM-R + \textit{TDA} & 73.0 \hsp 0.336 & 70.3 \hsp 0.297  & \underline{86.9} \hsp \underline{0.683} & 80.4 \hsp 0.522 \\

\midrule
\multicolumn{5}{c}{\textbf{ItaCoLA}} \\
\midrule

It-BERT  & 81.1 \hsp 0.032  & 82.1 \hsp 0.140 & 87.4 \hsp 0.351 & \textbf{86.8} \hsp 0.382 \\
It-BERT + $H0M$  & 85.0 \hsp 0.124 & 83.6 \hsp 0.055 & 87.0 \hsp 0.361 & 85.7 \hsp 0.370 \\
It-BERT + \textit{TDA} & \textbf{89.2} \hsp \textbf{0.478} & \textbf{85.8} \hsp \textbf{0.352} & \underline{91.1} \hsp \underline{0.597} & 86.4 \hsp \underline{0.424} \\

XLM-R  & 85.4 \hsp 0.000 & 84.2 \hsp 0.000 & 85.7 \hsp 0.397 & 85.6 \hsp \textbf{0.434} \\
XLM-R + $H0M$  & 84.7 \hsp 0.095 & 83.3 \hsp 0.072 & 86.9 \hsp 0.370 & \underline{86.6} \hsp 0.397 \\
XLM-R + \textit{TDA} & \underline{88.3} \hsp \underline{0.411} & \underline{84.4} \hsp \underline{0.208}  & \textbf{92.8} \hsp \textbf{0.683} & 86.1 \hsp 0.398 \\

\midrule
\multicolumn{5}{c}{\textbf{DaLAJ}} \\
\midrule

Sw-BERT & 58.3 \hsp 0.183 & 59.0 \hsp 0.188 & \underline{71.9} \hsp \underline{0.462} & \textbf{74.2} \hsp \textbf{0.500} \\
Sw-BERT + $H0M$   & \textbf{69.3} \hsp \textbf{0.387} & 58.4 \hsp 0.169 & \textbf{76.9} \hsp \textbf{0.542} & 68.7 \hsp 0.375 \\
Sw-BERT + \textit{TDA}  & \underline{62.1} \hsp \underline{0.243} & \textbf{64.4} \hsp \textbf{0.289} & 71.8 \hsp 0.442 & \underline{73.4} \hsp \underline{0.478} \\

XLM-R & 52.2 \hsp 0.069 & 51.5 \hsp 0.038 & 60.6 \hsp 0.243 & 62.8 \hsp 0.297 \\
XLM-R + $H0M$  & 61.1 \hsp 0.224 & \underline{61.8} \hsp \underline{0.237} & 62.5 \hsp 0.256 & 64.5 \hsp 0.295 \\
XLM-R + \textit{TDA} & 51.1 \hsp 0.227 & 54.1 \hsp 0.218 & 62.5 \hsp 0.255 & 65.5 \hsp 0.322 \\

\bottomrule
\end{tabular}
\caption{Acceptability classification results by benchmark. \textbf{IDD}=``in domain dev'' set (\textsc{CoLA}). \textbf{OODD}=``out of domain dev'' set (\textsc{CoLA}). \textbf{Dev}; \textbf{Test}=dev and test sets in \textsc{ItaCoLA} and \textsc{DaLAJ}. The best score is put in bold, the second best score is underlined.}

\label{table:cola_results}
\end{table}

\autoref{table:cola_results} outlines the LA classification results. Our \textit{TDA} classifiers generally outperform the baselines by up to $0.14$ MCC for English, $0.24$ MCC for Italian, and $0.08$ MCC for Swedish. The $H_0M$ feature can solely enhance the performance for English and Italian up to $0.1$, and concatenation of all features receives the best scores. Comparing the results under the \emph{frozen} and \emph{fine-tuned} settings, we draw the following conclusions. The \textit{TDA} features significantly improve the \emph{frozen} baseline performance but require the LM to be \emph{fine-tuned} for maximizing the performance. However, the \textit{TDA}/$H_0M$ classifiers perform on par with the \emph{fine-tuned} baselines for Swedish. The results suggest that our features may fail to infer lexical items and word derivation violations peculiar to the \textsc{DaLAJ} benchmark.

\begin{figure}[th!]
    \centering
    \includegraphics[width=0.8\linewidth]{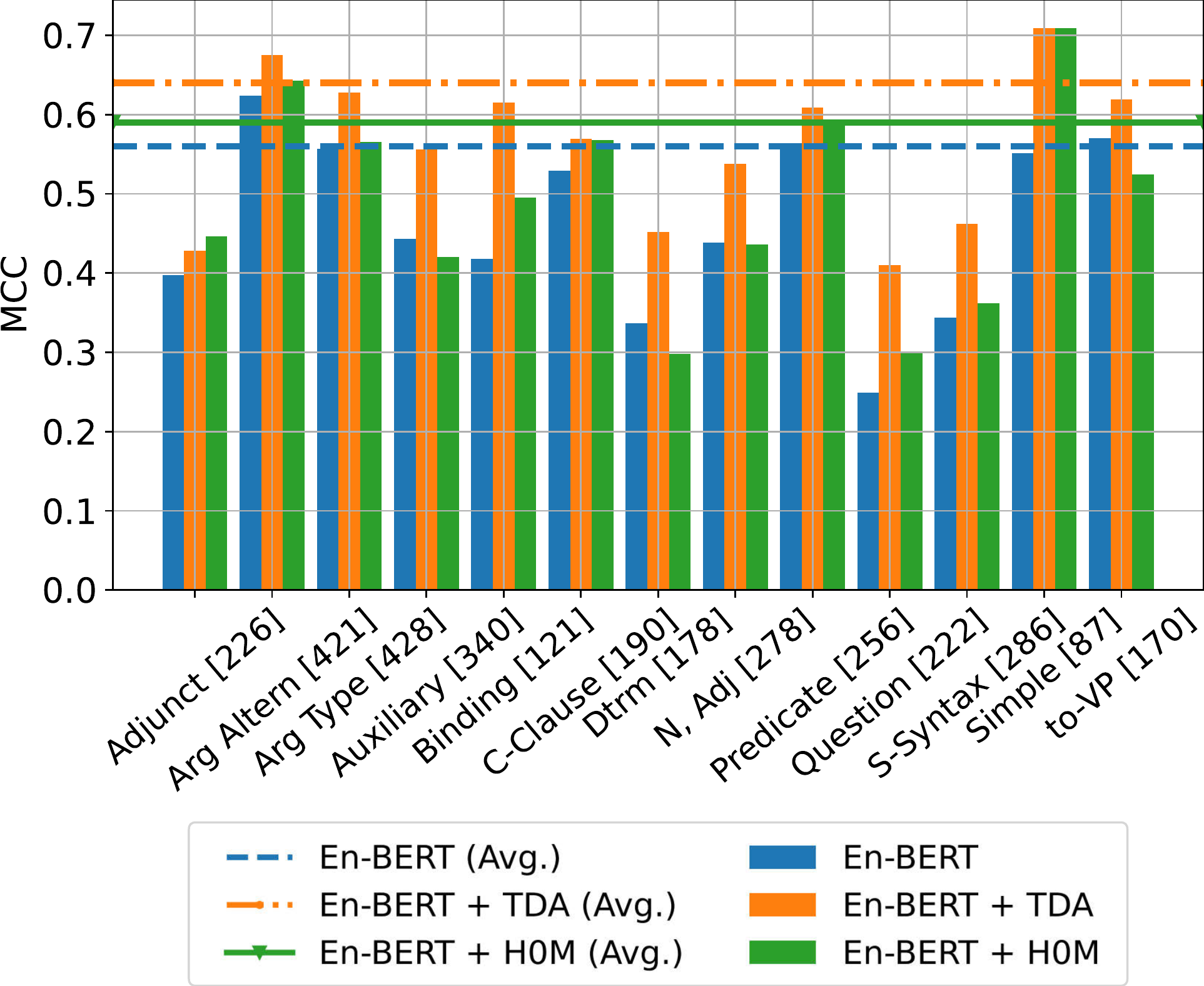}
    \caption{Performance (MCC) of the \emph{fine-tuned} En-BERT and XLM-R by major linguistic feature. Average MCC scores are represented with dashed lines. The number of sentences including the feature is placed in square brackets.}
    \label{fig:cola}
\end{figure}

\paragraph{Effect of Freezing Layers.} Another finding is that freezing the Transformer layers significantly affects acceptability classification. Most of the \emph{frozen} baselines score less than $0.1$ MCC across all languages. The results align with~\citet{lee2019would}, who discuss the performance degradation of BERT-based models depending upon the number of frozen layers. With all layers frozen, the model performance can fall to zero.

\paragraph{Results by Linguistic Features.} We run a diagnostic evaluation of the \emph{fine-tuned} models using a grammatically annotated version of the \textsc{CoLA} development set~\cite{warstadt2019linguistic}. \autoref{fig:cola} (En-BERT and XLM-R; \autoref{fig:cola_xlmr} in Appendix~\ref{sec:appendix_res_ling}) present the results of measuring the MCC of the sentences including the major features.

The overall pattern is that the \textit{TDA} classifiers may outperform the \emph{fine-tuned} baselines, while the $H_0M$ ones perform on par with the latter. The performance is high on sentences with default syntax (\textit{Simple}) and marked argument structure, including prepositional phrase arguments (\textit{Arg. Type}), and verb phrases with unusual structures (\textit{Arg. Altern}). The \textit{TDA} features capture surface properties, such as presence of auxiliary or modal verbs (\textit{Auxiliary}), and structural ones, e.g., embedded complement clauses (\textit{Comp Clause}) and infinitive constructions (\textit{to-VP}). The models receive moderate MCC scores on sentences with question-like properties (\textit{Question)}, adjuncts performing semantic functions (\textit{Adjunct}), negative polarity items, and comparative constructions (\textit{Determiner}).

\paragraph{Analysis of the Feature Space.}
The LA classification experiments are conducted in the sparse feature space, where the \textit{TDA} features can strongly correlate with one another, and their contribution is unclear. We run a complementary experiment to understand better how linguistic features are modeled with topology. We investigate the feature space with dimensionality reduction (principal component analysis, PCA;~\citealp{pearson1901liii}) by interpreting components' structure and identifying the feature importance to the classifier's predictions using Shapley values~\cite{shapley1953value}, a game-theoretic approach to the attribution problem~\cite{sundararajan2020many}. Appendix~\ref{sec:appendix_pca} describes the experiment on the \textit{fine-tuned} En-BERT + \textit{TDA} model using the grammatically annotated \textsc{CoLA} development set. 

The results show that \emph{(i)}~features of the higher-layer heads, such as the average vertex degree, the number of connected components, edges, and cycles, and attention to the current token, contribute most to the major linguistic features. \emph{(ii)}~Attention to the \texttt{[CLS]}/next token is important to the \textit{Determiner}, \textit{Arg. Type}, \textit{Comp Clause}, and \textit{to-VP} properties, while attention to the first token and punctuation marks has the least effect in general. \emph{(iii)}~The number of nodes influences the classifier behavior, which is in line with~\citeauthor{warstadt2019linguistic}, who discuss the effect of the sentence length on the performance.

\section{Linguistic Minimal Pairs}
\label{sec:minimal_pairs}

\subsection{Data}
\textsc{BLiMP}~(Benchmark of Linguistic Minimal Pairs;~\citealp{warstadt-etal-2020-blimp-benchmark}) evaluates the sensitivity of LMs to acceptability contrasts in terms of a forced choice between minimal pairs, as in Example~\ref{eg:blimp}. The benchmark consists of $67$ pair types, each including $1$k pairs covering $12$ language phenomena in morphology, syntax, and semantics.

\decreasespace
\ex. \label{eg:blimp} 
\a. Whose hat should Tonya wear?
\b. *Whose should Tonya wear hat?

\subsection{Models}
We conduct the experiments using two Transformer LMs for English: BERT-base and RoBERTa-base~\cite{liu2019roberta}. 

\paragraph{Baselines.} We compare our methods with the results on \textsc{BLiMP} for human annotators and nine LMs~\cite{warstadt-etal-2020-blimp-benchmark,salazar-etal-2020-masked}. The baselines range from statistical N-gram LMs to Transformer LMs.

\paragraph{Our Models.} 
Given a minimal pair as in Example~\ref{eg:rtd_pair}, we build attention graphs $G_a$ and $G_b$ from each attention head of a \textit{frozen} Transformer LM. We use the $H_0M$ feature (\S\ref{persistent_features}) and RTD (\S\ref{rtd_method}) as scoring functions to distinguish between the sentences $S_a$ and $S_b$. The scoring is based on empirically defined decision rules modeled after the forced-choice task:

\begin{itemize}[noitemsep]
    \item $H_0M$ scoring: $S_a$ is acceptable if and only if $H_0M(G_a)~<~H_0M(G_b)$; otherwise $S_b$ is acceptable. 
    
    \item RTD scoring: $S_a$ is acceptable if and only if RTD$(G_a,~G_b)~<~$RTD$(G_b,~G_a)$; otherwise $S_b$ is acceptable.
\end{itemize}

\noindent We evaluate the scoring performance of each attention head, head ensembles, and all heads w.r.t. each and all linguistic phenomena in \textsc{BLiMP}. The following head configurations are used for each Transformer LM and scoring function:

\begin{table*}[!ht]
\centering
\small
\resizebox{0.96\textwidth}{!}{
\begin{tabular}{l | c | cccccccccccc} 
\toprule

\textbf{Model} & \rotatebox{10}{\textbf{Overall}} & \rotatebox{10}{\textbf{Ana. agr}} & \rotatebox{10}{\textbf{Arg. str}} & \rotatebox{10}{\textbf{Binding}} & \rotatebox{10}{\textbf{Ctrl./rais.}} & \rotatebox{10}{\textbf{D-N agr}} & \rotatebox{10}{\textbf{Ellipsis}} & \rotatebox{10}{\textbf{Filler}} & \rotatebox{10}{\textbf{Irreg.}} & \rotatebox{10}{\textbf{Island}} & \rotatebox{10}{\textbf{NPI}} & \rotatebox{10}{\textbf{Quant.}} & \rotatebox{10}{\textbf{S-V agr}} \\

\midrule
\multicolumn{14}{l}{\citet{warstadt-etal-2020-blimp-benchmark}} \\ 
\midrule

$5$-gram & 61.2 & 47.9 & 71.9 & 64.4 & 68.5 & 70.0 & 36.9 & 60.2 & 79.5 & 57.2 & 45.5 & 53.5 & 60.3 \\
LSTM & 69.8 & 91.7 & 73.2 & 73.5 & 67.0 & 85.4 & 67.6 & 73.9 & 89.1 & 46.6 & 51.7 & 64.5 & 80.1 \\
Transformer-XL & 69.6 & 94.1 & 72.2 & 74.7 & 71.5 & 83.0 & 77.2 & 66.6 & 78.2 & 48.4 & 55.2 & 69.3 & 76.0 \\
GPT2-large & 81.5 & \textbf{99.6} & 78.3 & 80.1 & 80.5 & 93.3 & 86.6 & 81.3 & 84.1 & 70.6 & 78.9 & 71.3 & 89.0 \\
{Human Baseline} & \underline{88.6} & {97.5} & \textbf{90.0} & {87.3} & \underline{83.9} & {92.2} & {85.0} & {86.9} & \textbf{97.0} & \textbf{84.9} & {88.1} & \textbf{86.6} & {90.9} \\

\midrule

\multicolumn{14}{l}{\citet{salazar-etal-2020-masked}}  \\ 
\midrule
GPT2-medium & 82.6 & \underline{99.4} & 83.4 & 77.8 & {83.0} & 96.3 & 86.3 & 81.3 & 94.9 & 71.7 & 74.7 & 74.1 & 88.3 \\
BERT-base & 84.2 & 97.0 & 80.0 & 82.3 & 79.6 & \textbf{97.6} & 89.4 & 83.1 & 96.5 & 73.6 & 84.7 & 71.2 & 92.4 \\
BERT-large & 84.8 & 97.2 & 80.7 & 82.0 & {82.7} & \textbf{97.6} & 86.4 & 84.3 & 92.8 & 77.0 & 83.4 & 72.8 & {91.9} \\
RoBERTa-base & 85.4 & 97.3 & {83.5} & 77.8 & 81.9 & \underline{97.0} & \textbf{91.4} & \underline{90.1} & 96.2 & 80.7 & 81.0 & 69.8 & {91.9} \\
RoBERTa-large & 86.5 & 97.8 & \underline{84.6} & 79.1 & \textbf{84.1} & 96.8 & 90.8 & 88.9 & \underline{96.8} & \underline{83.4} & 85.5 & 70.2 & 91.4 \\

\midrule

\multicolumn{14}{l}{BERT-base + $H_0M$}  \\
\midrule

\scriptsize{[Layer; Head]} & \scriptsize{\xmark} &  \scriptsize{[8; 8]} &  \scriptsize{[8; 0]} &  \scriptsize{[7; 0]}&  \scriptsize{[8; 0]} &  \scriptsize{[7; 0]}&  \scriptsize{[7; 11]} &   \scriptsize{[9; 7]} &  \scriptsize{[6; 1]} &  \scriptsize{[11; 7]} &  \scriptsize{[8; 9]} &  \scriptsize{[3; 7]} &  \scriptsize{[8; 0]}  \\
Phenomenon Head   & 81.7 & 94.9 & 75.9 & 80.4 & 79.2 & 96.7 & 89.1 &  75.9 & 93.0 & 70.5 & 84.6 & 81.2 & 82.1  \\

Top Head \scriptsize{[8; 0]}   & 75.4 & {86.8} & {75.9} & {63.4} & {79.2} & {83.7} & {72.2} &  {67.3} & {90.3} & {70.0} & {83.1} & {63.5} & {82.1}  \\

Head Ensemble   & {84.3} & {93.3} & {79.9} & {83.5} & {78.6} & {96.4} & {78.4} & {79.5} &  {93.8} & {74.4} & {92.5} & {81.7} & {86.8}  \\

All Heads & {64.8} & {79.6} & {69.1} & {63.9} & {62.6} & {86.2} & {70.7} &  {47.3} & {90.7} & {49.5} & {61.1} & {50.0} & {72.0}  \\

\midrule
\multicolumn{14}{l}{BERT-base + RTD} \\ 
\midrule

\scriptsize{[Layer; Head]} &  \scriptsize{\xmark}  &\scriptsize{[8; 3]} & \scriptsize{[8; 0]} & \scriptsize{[7; 0]} &  \scriptsize{[8; 0]} & \scriptsize{[7; 0]} & \scriptsize{[7; 11]} &  \scriptsize{[9; 7]} & \scriptsize{[6; 1]} & \scriptsize{[9; 7]} & \scriptsize{[8; 9]} & \scriptsize{[3; 7]} & \scriptsize{[8; 0]}  \\

Phenomenon Head & {81.8} & {94.5} & {75.8} & {80.4} & {79.2} & {96.7} & {89.1} &  {75.0} & {93.0} & {72.2} & {84.4} & {81.2} & {82.0}  \\

Top Head  \scriptsize{[8; 0]}  & {75.4} & {86.8} & {75.8} & {63.3} & {79.2} & {83.6} & {72.1} &  {67.3} & {90.2} & {70.2} & {83.1} & {63.6} & {82.0}  \\

Head Ensemble   & 85.8 & 93.9 & 82.5 & 85.6 & 77.0 & 96.3 & 88.1 & 80.7 & 95.7 & 77.0 & 92.5 & {83.8} & 88.9 \\
All Heads & {65.3} & {77.8} & {68.5} & {63.2} & {63.6} & {86.0} & {73.4} &  {48.4} & {91.0} & {51.3} & {62.0} & {51.2} & {72.6}  \\

\midrule

\multicolumn{14}{l}{RoBERTa-base + $H_0M$}  \\
\midrule

\scriptsize{[Layer; Head]} &   \scriptsize{\xmark} & \scriptsize{[9; 5]}  &  \scriptsize{[9; 8]} &  \scriptsize{[9; 5]}&  \scriptsize{[9; 6]} &  \scriptsize{[9; 0]} &  \scriptsize{[8; 4]} &   \scriptsize{[11; 10]} &  \scriptsize{[9; 6]} &  \scriptsize{[3; 5]} &  \scriptsize{[11; 5]} &  \scriptsize{[11; 3]} &  \scriptsize{[8; 9]}  \\

Phenomenon Head & {86.5} & {97.6} & {79.9} & \textbf{90.5} & {80.7} & {91.6} & {89.9} &  {87.1} & {95.9} & {78.9} & {91.1} & {83.4} & {90.2}  \\

Top Head  \scriptsize{[11; 10]} & {81.9} & {90.1} & {66.0} & {84.7} & {71.7} & {91.0} & {86.7} & {87.1} &  {89.5} & {76.8} & {90.7} & {78.4} & {85.0}  \\

Head Ensemble   & {87.8} & {96.3} & {79.6} & {87.6} & {82.6} & {93.6} & {84.9} & \textbf{90.4} &  {94.3} & {83.0} & \textbf{94.6} & {80.6} & \textbf{92.8}  \\
All Heads & {74.3} & {80.6} & {71.2} & {78.6} & {67.8} & {90.6} & {89.9} &  {75.5} & {65.1} & {73.4} & {65.1} & {57.0} & {75.6}  \\

\midrule
\multicolumn{14}{l}{RoBERTa-base + RTD} \\ 

\midrule

\scriptsize{[Layer; Head]} &   \scriptsize{\xmark} & \scriptsize{[9; 5]}  &  \scriptsize{[9; 8]} &  \scriptsize{[9; 5]}&  \scriptsize{[9; 6]} &  \scriptsize{[9; 0]} &  \scriptsize{[8; 4]} &   \scriptsize{[11; 10]} &  \scriptsize{[2; 9]} &  \scriptsize{[3; 5]} &  \scriptsize{[10; 2]} &  \scriptsize{[11; 3]} &  \scriptsize{[9; 5]}  \\

Phenomenon Head & {86.8} & {97.6} & {80.7} & \underline{90.3} & {80.7} & {91.9} & {90.4} &  {86.8} & {95.9} & {78.9} & {91.3} & {82.9} & {90.4}  \\

Top Head \scriptsize{[11; 10]} & {80.8} & {88.8} & {63.3} & {83.7} & {69.8} & {90.2} & {88.8} & {86.8} &  {89.1} & {76.5} & {89.0} & {78.4} & {83.1}  \\

Head Ensemble   & \textbf{88.9} & {97.0} & {83.3} & {86.9} & {81.6} & {93.7} & {88.1} & {88.8} &  {95.5} & {83.0} & \underline{94.2} & \underline{84.2} & \underline{92.7}  \\

All Heads & {74.3} & {81.1} & {66.6} & {76.8} & {65.2} & {89.9} & \underline{91.1} &  {74.6} & {63.3} & {74.1} & {63.2} & {56.9} & {74.9}  \\

\bottomrule
\end{tabular}}
\caption{Percentage accuracy of the baseline models, human baseline, and our methods on \textsc{BLiMP}. \textbf{Overall} is the average across all phenomena. The best score is put in bold, the second best score is underlined.}
\label{table:blimp_results}
\end{table*}

\begin{itemize}[noitemsep]
    \item \textit{Phenomenon Head} and \textit{Top Head}  are the best-performing attention heads for each and all phenomena, respectively. The heads undergo the selection with a brute force search and operate as independent scorers.
    
    \item \textit{Head Ensemble} is a group of the best-performing attention heads selected with beam search. The size of the group is always odd. We collect majority vote scores from attention heads in the group.
    
    \item \textit{All Heads} involves majority vote scoring with all $144$ heads. We use random guessing in case of equality of votes. This setup serves as a proxy for the efficiency of the head selection.
\end{itemize}

\paragraph{Notes on Head Selection.} Recall that the head selection procedure\footnote{We acknowledge that this setup may contradict the original task formulation, that is, evaluation of pre-trained LMs on minimal pairs in the unsupervised setting~\cite{warstadt-etal-2020-blimp-benchmark}. We use auxiliary data to find the most contributing heads and analyze their linguistic roles via TDA \emph{without} any model parameter updates. The generated data is publicly available for reproducibility purposes.} imposes the following limitation. Auxiliary labeled minimal pairs are required to find the best-performing Phenomenon, Top Heads, and Head Ensembles. However, this procedure is more optimal and beneficial than All Heads since it maximizes the performance when utilizing only one or $9$-to-$59$ heads. We also analyze the effect of the amount of auxiliary data used for the head selection on the scoring performance (\S\ref{subsection:blimp_results}). Appendix~\ref{subsection:head_selection} presents a more detailed description of the head selection procedure.

\subsection{Results}
\label{subsection:blimp_results}
We provide the results of scoring \textsc{BLiMP} pairs in~\autoref{table:blimp_results}. The accuracy is the proportion of the minimal pairs in which the method prefers an acceptable sentence to an unacceptable one. We report the maximum accuracy scores for our methods across five experiment restarts. The general trends are that the best head configuration performs on par with the human baseline and achieves the highest overall performance (RoBERTa-base + RTD; Head Ensemble). RoBERTa predominantly surpasses BERT and other baselines, and topological scoring may improve on scores from both BERT and RoBERTa for particular phenomena.

\paragraph{Top Head Results.}
We find that $H_0M$/RTD scoring with only one Top Head overall outperforms majority voting of $144$ heads (All Heads) by up to $10.6$\% and multiple baselines by up to $20.7$\% ($5$-gram, LSTM, Transformer-XL, GPT2-large). However, this head configuration performs worse than masked LM scoring~\cite{salazar-etal-2020-masked} for BERT-base (by $8.8$\%; Top Head=[$8$; $0$]) and RoBERTa-base (by $4.6$\%; Top Head=[$11$; $10$]). 

\paragraph{Phenomenon Head Results.}
We observe that the $H_0M$/TDA scoring performance of Phenomenon Heads insignificantly differ for the same model. Phenomenon Heads generally receive higher scores than the corresponding Top Heads for BERT/RoBERTa~(e.g.,~\textit{Binding}:~+$17.1$/+$5.8$\%;~\textit{Quantifiers}:~+$17.7$/+$5.0$\%;~\textit{Det.-Noun agr}:~+$13.0$/+$1.7$\%), and perform best or second-best on \textit{Binding} and \textit{Ellipsis}. Their overall performance further adds up to $6.4$/$6.0$\% and is comparable with~\citeauthor{salazar-etal-2020-masked}. The results indicate that heads encoding the considered phenomena are distributed at the same or nearby layers, namely $[3;$ $6$-$9; 11]$ (BERT), and $[2$-$3; 8$-$11]$ (RoBERTa).

\begin{table}[t!]
    \centering
    \scriptsize
    \begin{tabular}{cc}
    \toprule
    {BERT-base} & {RoBERTa-base} \\ \midrule
             
    \includegraphics[width=0.2\textwidth]{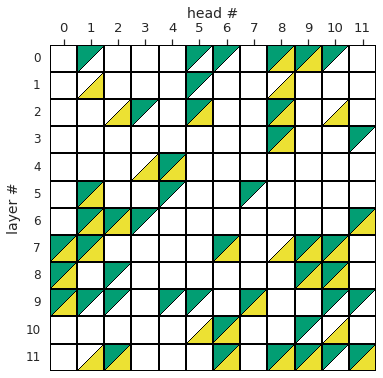} &
    \includegraphics[width=0.2\textwidth]{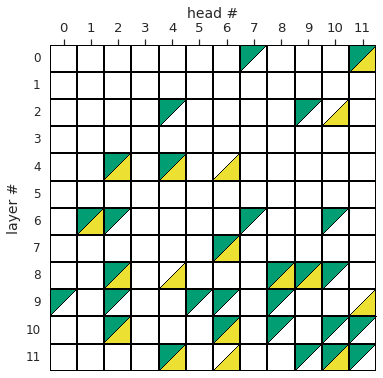} \\
             
    \bottomrule
\end{tabular}
\caption{Results of selecting the best-performing Head Ensembles with $H_0M$/RTD-based scoring. $H_0M$ heads are colored in green; RTD heads are colored in yellow.}
\label{tab:voting_results}
\end{table}

\paragraph{Head Ensemble Results.}
\autoref{tab:voting_results} describes the most optimal Head Ensembles by Transformer LM. Most heads selected under the $H_0M$ and RTD scoring functions are similar w.r.t. LMs. While the selected BERT heads are distributed across all layers, the RoBERTa ones tend to be localized at the middle-to-higher layers. Although RoBERTa utilizes smaller ensembles when delivering the best overall score, some heads contribute in both LMs, most notably at the higher layers. 

Overall, the RoBERTa $H_0M$/RTD ensembles get the best results on \textit{Filler gap}, \textit{Quantifiers}, \textit{Island effects}, \textit{NPI}, and \textit{S-V agr} as shown in~\autoref{table:blimp_results}, matching the human level and surpassing four larger LMs on all phenomena by up to $7.4$\% (GPT2-medium and GPT2/BERT/RoBERTa-large).

\paragraph{Effect of Auxiliary Data.} Note that the head selection can be sensitive to the number of additional examples. The analysis of this effect is presented in Appendix~\ref{subsection:auxdata}. The results show that head ensembles, their size, and average performance tend to be more stable when using sufficient examples (the more, the better); however, using only one extra example can yield the performance above $80$\%.

\section{Discussion}
\label{sec:discussion}
\begin{figure}[t!]
  \centering
  \includegraphics[width=0.95\linewidth]{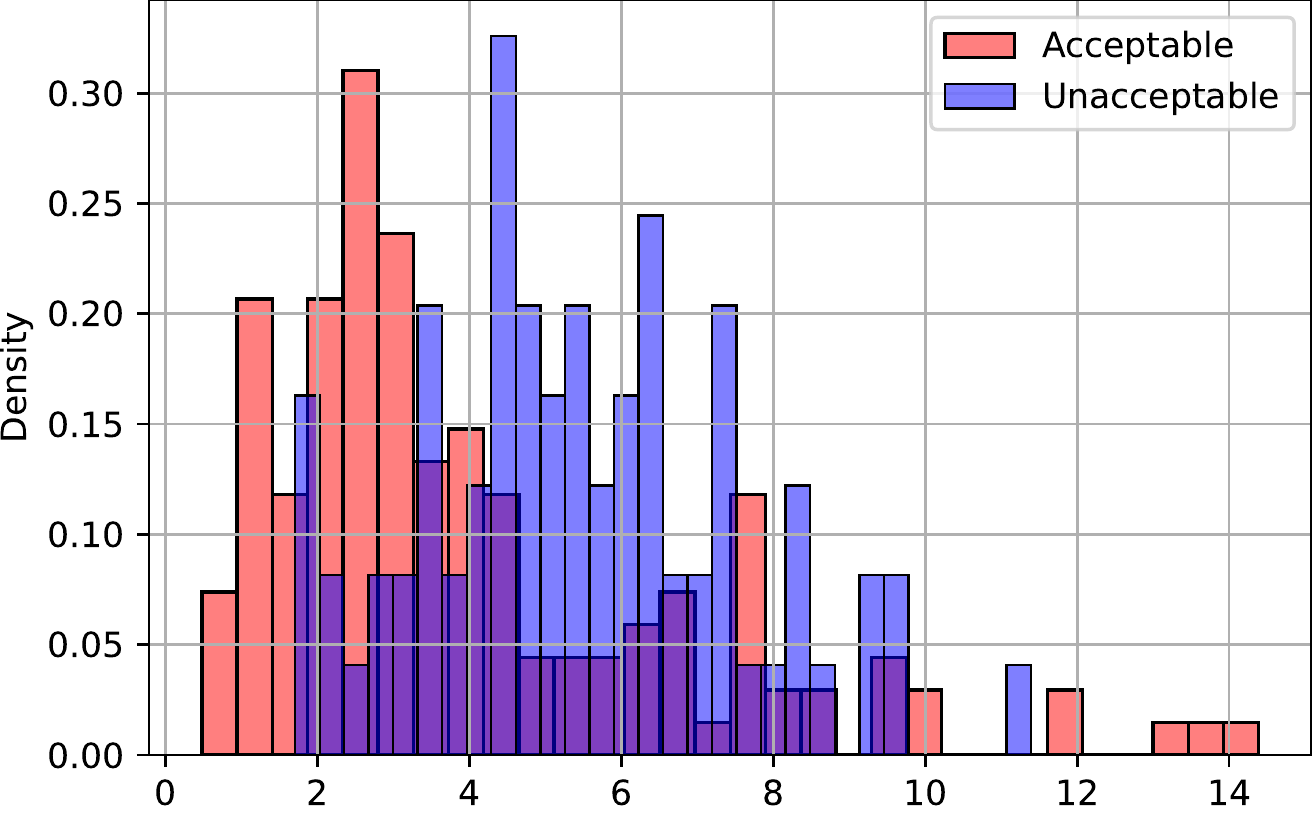}
  \caption{The distribution shift of the $H_0S$ feature between the acceptable and unacceptable sentences (\textit{Question}); [L: $10$; H: $0$].}
  \label{fig:h0s_question}
\end{figure}

\noindent \textbf{Topology and Acceptability.} The topological properties of the attention graph represent interpretable and versatile features for judging sentence acceptability and identifying acceptability contrasts in minimal pairs. As one of such properties, the sum length of bars ($H_0S$) --- and its normalized version ($H_0M$) --- have proved to be efficient for both LA approaches. This simple feature can serve as a profitable input for LA classifiers and a scoring function to discriminate between minimal pairs. \autoref{fig:h0s_question} shows an example of the $H_0S$ sensitivity to \textsc{CoLA}'s question-like properties, such as wh-movement out of syntactic islands and matrix and embedded questions. We provide more examples in \autoref{appendix:h0s}, which demonstrate the distribution shifts between the acceptable and unacceptable sentences.

\paragraph{Acceptability Phenomena.} The underlying structure of the attention graph encodes various well-established grammatical concepts. We observe that the persistent graph features capture surface properties, morphological agreement, structural relationships, and simple/complex syntactic phenomena well. However, with topology, lexical items, optional syntactic elements, and abstract semantic factors may be difficult to infer. Attention to the first token and punctuation marks contribute least to LA classification, while the other attention pattern features capture various phenomena.

\paragraph{Linguistic Roles of Heads.} Topological tools help gain empirical evidence about the linguistic roles of heads from another perspective. Our findings on the heads' roles align with several related studies. The results on the \textsc{CoLA}-style and \textsc{BLiMP} benchmarks indicate that \emph{(i)}~a single head can perform multiple linguistic functions~\cite{pande2021heads}, \emph{(ii)}~some linguistic phenomena, e.g., phrasal movement and island effects, are better captured by head ensembles rather than one head~\cite{htut2019attention}, and \emph{(iii)}~heads within the same or nearby layers extract similar grammatical phenomena~\cite{bian-etal-2021-attention}.

\section{Conclusion and Future Work}
Our paper studies the ability of attention heads to judge grammatical acceptability, demonstrating the profitable application of TDA tools to two LA paradigms. Topological features can boost LA classification performance in three typologically close languages. The $H_0M$/RTD scoring matches or outperforms larger Transformer LMs and reaches human-level performance on \textsc{BLiMP}, while utilizing $9$-to-$59$ attention heads. We also interpret the correspondence between the persistent features of the attention graph and grammatical concepts, revealing that the former efficiently infer morphological, structural, and syntactic phenomena but may lack lexical and semantic information. 

In our future work we hope to assess the linguistic competence of Transformer LMs on related resources for typologically diverse languages and analyze which language-specific phenomena are and are not captured by the topological features. We are also planning to examine novel features, e.g., the number of vertex covers, the graph clique-width, and the features of path homology~\cite{grigor2020path}. Another direction is to evaluate the benefits and limitations of the $H_0M$/RTD features as scoring functions in downstream applications.

We also plan to introduce support for new deep learning frameworks such as MindSpore\footnote{https://www.mindspore.cn/} \cite{9701700} to bring TDA-based experimentation to the wider industrial community.

\section{Limitations}
\subsection{Computational complexity} 
\noindent\textbf{Acceptability classification.} Calculation of any \textbf{topological feature} relies on the Transformer's attention matrices. Hence, the computational complexity of our features is not lower than producing an attention matrix with one head, which is asymptotically $O(n^2d + nd^2)$ given that $n$ is the maximum number of tokens, and $d$ is the token embedding dimension~\cite{vaswani2017attention}.

The calculation complexity of the \textbf{pattern-based} and \textbf{threshold-based features} is done in linear time $O(e + n)$, where $e$ is the number of edges in the attention graph. In turn, the number of edges is not higher than $\frac{n(n-1)}{2} \sim n^2$. The computation of \textbf{the 0$^{th}$ Betti number} $\beta_0$ takes linear time $O(e + n)$, as $\beta_0$ is equal to the number of the connected components in an undirected graph. The computation of \textbf{the 1$^{st}$ Betti number} $\beta_1$ takes constant time, since $\beta_1 = e - n + \beta_0$. The computational complexity of \textbf{the number of simple cycles} and the \textbf{$1$-dimensional barcode features} is exponential in the worst case. To reduce the computational burden, we stop searching for simple cycles after a pre-defined amount of them is found.

\begin{figure}[t!]
  \centering
  \includegraphics[width=\linewidth]{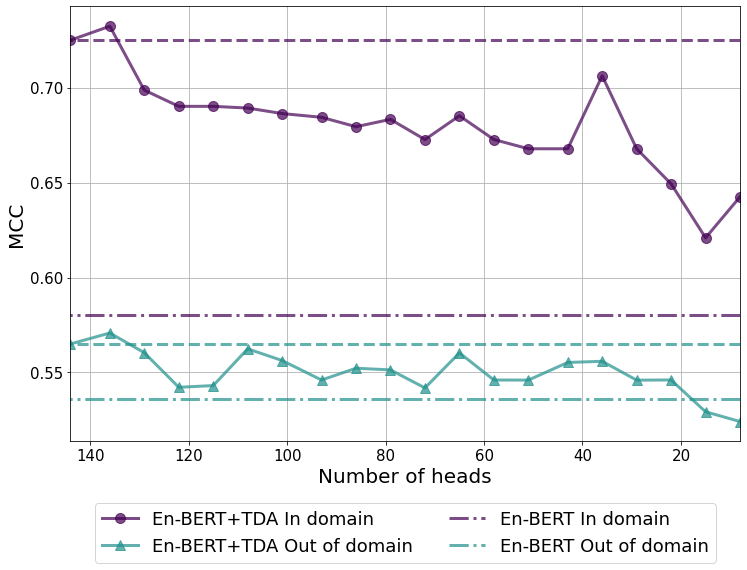}
  \caption{Performance on the \textsc{CoLA} development set depending on the number of heads for En-BERT + \textit{TDA}.}
  \label{fig:mcc_heads}
\end{figure}

Note that the computational costs could be reduced, e.g., by identifying the most contributing features or the best-performing heads. Consider an example in~\autoref{fig:mcc_heads}, which illustrates how the \textsc{CoLA} performance gets changed depending on the number of En-BERT heads. Here, the head selection is based on a simple procedure. First, we score the attention heads by calculating the maximum correlation between each head's features and the vector of the target classes on the train set. Second, we train a linear classifier over the \textit{TDA} features produced by $N$ attention heads ranked by the correlation values, as specified in \S\ref{la_models}. Satisfactory MCC scores can be achieved when utilizing less than $40$ heads with a significant speed up at the inference stage.

\vspace{0.2em} \noindent\textbf{Linguistic Minimal Pairs.} Computation of the $H_0M$ and RTD features is run via the Ripser++ GPU library. Under this library, the minimum spanning tree is found according to Kruskal's algorithm giving the computational complexity of $H_0M$ as $O(n^2\log n)$. The complexity can be reduced using other algorithms, e.g., Prim's algorithm, which takes $O(n^2)$. The RTD's computational complexity is more difficult to estimate. RTD is computed via persistence barcodes of dimension 1 for a specific graph with $2n$ vertices. Many optimization techniques and heuristics are implemented in the Ripser++ library that significantly reduce the RTD's complexity.

\paragraph{Empirical estimate.} Computing the $H_0M$/RTD features with $144$ BERT heads in the worst case of a $512$-token text takes $2.41$ and $94.5$ sec (NVIDIA Tesla K80 $12$GB RAM). However, the actual computation time on the considered tasks is empirically more optimal. We provide empirical estimates on the entire BLiMP and LA datasets: $2.4$/$15.7$ hours on BLiMP ($H_0M$/RTD) and up to $2$ hours on \textsc{CoLA}/\textsc{ItaCoLA}/\textsc{DaLAJ} (estimates by the feature groups: topological features=$24$\%; features derived from barcodes=$70$\%; and features based on distance to patterns=$6$\% of the total time).

\vspace{0.2em} \noindent \subsection{Application Limitations} We also outline several application limitations of our approach. \emph{(i)} The LA classifiers require preliminary fine-tuning of Transformer LMs to extract more representative attention graph features and, therefore, achieve better performance. \emph{(ii)} RTD operates upon a one-to-one vertex correspondence, which may be hindered by tokens segmented into an unequal amount of sub-tokens. As a result, identifying the topological discrepancy between pairs of attention graphs can be restricted in practice, where the graphs are of an arbitrary number of nodes. Regardless of the potential information loss due to sentence truncation in such cases, the RTD heads still receive the best overall score on \textsc{BLiMP}. \emph{(iii)} The head selection procedure relies on auxiliary data to identify the best-performing head configurations. Annotating the auxiliary data may require additional resources and expertise for practical purposes. However, the procedure maximizes the performance and reduces the computational costs by utilizing less attention heads.

\subsection{Linguistic Acceptability}
Acceptability judgments have been broadly used to investigate whether LMs learn grammatical concepts central to human linguistic competence. However, this approach has several methodological limitations. \emph{(i)} The judgments may display low reproducibility in multiple languages~\cite{linzen2018reliability}, and \emph{(ii)} be influenced by an individual's exposure to ungrammatical language use~\cite{dkabrowska2010naive}. \emph{(iii)} Distribution shifts between LMs' pre-training corpora and LA datasets may introduce bias in the evaluation since LMs tend to assign higher probabilities to frequent patterns and treat them as acceptable in contrast to rare ones~\cite{marvin-linzen-2018-targeted,linzen2021syntactic}.

\section{Ethical Statement} 
Advancing acceptability evaluation methods can improve the quality of natural language generation~\cite{batra-etal-2021-building}. We recognize that this, in turn, can increase the misuse potential of such models, e.g., generating fake product reviews, social media posts, and other targeted manipulation~\cite{jawahar-etal-2020-automatic,weidinger2021ethical}. However, the acceptability classifiers and scoring functions laid out in this paper are developed for research purposes only. Recall that the topological tools can be employed to develop adversarial defense and artificial text detection models for mitigating the risks~\cite{kushnareva-etal-2021-artificial}.

\section*{Acknowledgements}
This work was supported by Ministry of Science and Higher Education grant No. 075-10-2021-068 and by the Mindspore community. Irina Proskurina was supported by the framework of the HSE University Basic Research Program.

\bibliography{anthology,custom}
\bibliographystyle{acl_natbib}

\appendix
\newpage
\clearpage

\setcounter{figure}{0}
\setcounter{table}{0}

\section{Representation Topology Divergence}
\label{sec:appendix_rtd}
Suppose we have two weighted full graphs $G_a, G_b$ with one-to-one vertex correspondence. Define their vertices as $\lbrace a_1, a_2, \ldots , a_n\rbrace$ and $\lbrace b_1, b_2, \ldots , b_n\rbrace$ respectively so that $a_i$ corresponds to $b_i$ for each $i$. RTD($G_a, G_b$) is calculated as follows:

\begin{enumerate}[noitemsep]
\item Build a full weighted graph $G_{ab}$ with the vertices set $V = \lbrace v_1, v_2, \ldots v_n, u_1, u_2, \ldots, u_n\rbrace$ and the edge weights computed as
\begin{equation*}
\begin{cases}
w(v_i, v_j) = 0 \\
w(v_i, u_i) = 0 \\
w(u_i, u_j) = w_b (b_i, b_j) \\
w(v_i, u_j) = \max{(w_a(a_i, a_j), w_b(b_i, b_j))} \\
\end{cases}
\end{equation*}
where $w_a$ and $w_b$ are the edge weights in the corresponding graphs.

\item Compute the barcode~\cite{barannikov2021canonical} of the $H_1$ homology group of the graph $G_{ab}$ flag complex. 
It should be emphasized that the $H_0$ homology group barcode for this graph is empty since the minimum spanning tree of $G_{ab}$ has the total weight of $0$. Instead of $H_1$, the higher-order homology groups (e.g., $H_2$, $H_3$) can be considered. However, the preliminary experiments have shown that they are less helpful for LA tasks.

\item RTD$(G_a, G_b)$ is calculated as the sum of bar lengths in the barcode from the previous step. 
\end{enumerate}

It should be noted that this procedure is asymmetric on $G_a$ and $G_b$, and for non-equal graphs holds RTD$(G_a, G_b) \neq$ RTD$(G_b, G_a)$. To compute barcodes, we use the Ripser++ toolkit, which cannot work with asymmetric graphs. Hence, we represent the asymmetric attention maps as the distance matrices to obtain the symmetric graphs $G_a$ and $G_b$ as described in \S\ref{persistent_features}. We consider only the forward-looking part of attention, i.e., how each token affects the rest of the sentence. 

The majority of the \textsc{BLiMP} minimal pairs are of equal length in the BERT/RoBERTa tokens. Otherwise, we truncate the longest sentence to achieve an equal length since the one-to-one correspondence between tokens is crucial for RTD. We assume that the truncation may remove tokens that help to discriminate between the acceptable and unacceptable sentences. We leave improvement of the pre-processing stage for future work.

\section{Fine-tuning Details}
\label{sec:appendix_training}
Fine-tuning and evaluation of the BERT-based/XLM-R acceptability classifiers follow the standard procedure under the HuggingFace library~\cite{wolf-etal-2020-transformers}. Each model is fine-tuned for $4$ epochs with the learning rate of $1e^{-2}$/$1e^{-3}$, batch size of $32$, and the other default hyperparameters.

\section{Acceptability Classification}
\label{sec:appendix_cola}
\begin{table}[h!]
\centering
\scriptsize
\setlength{\tabcolsep}{2pt}
\begin{tabular}{lrrr} 
\toprule
 & \textbf{CoLA} & \textbf{ItaCoLA} & \textbf{DaLAJ} \\ 
\midrule
\textbf{Language} & English & Italian & Swedish \\
\textbf{\# Train sent.}  & 8,551 & 7,801 & 6,870 \\
\textbf{\# Dev sent.}  & 1,043 & 946 & 892 \\
\textbf{\# Test sent.} & 1,063 & 975 & 952 \\
\textbf{Type} & Expert & Expert & L2 \\
\textbf{\# Sources} & 23 & 12 & SweLL \\
\textbf{Phenomena} & Morph, Syntax, Semantics & Syntax & Lexis, Morph \\
\textbf{\%} & 70.5 & 84.5 & 50.0 \\ 

\bottomrule
\end{tabular}
\caption{Statistics of acceptability classification benchmarks. \textbf{Type}=Type of data source. \textbf{\%}=Percentage of acceptable sentences. \textbf{Morph}=Morphology.}
\label{table:data_stats}
\end{table}

\subsection{Results by Linguistic Features}
\label{sec:appendix_res_ling}

\begin{figure}[h!]
    \centering
    \includegraphics[width=\linewidth]{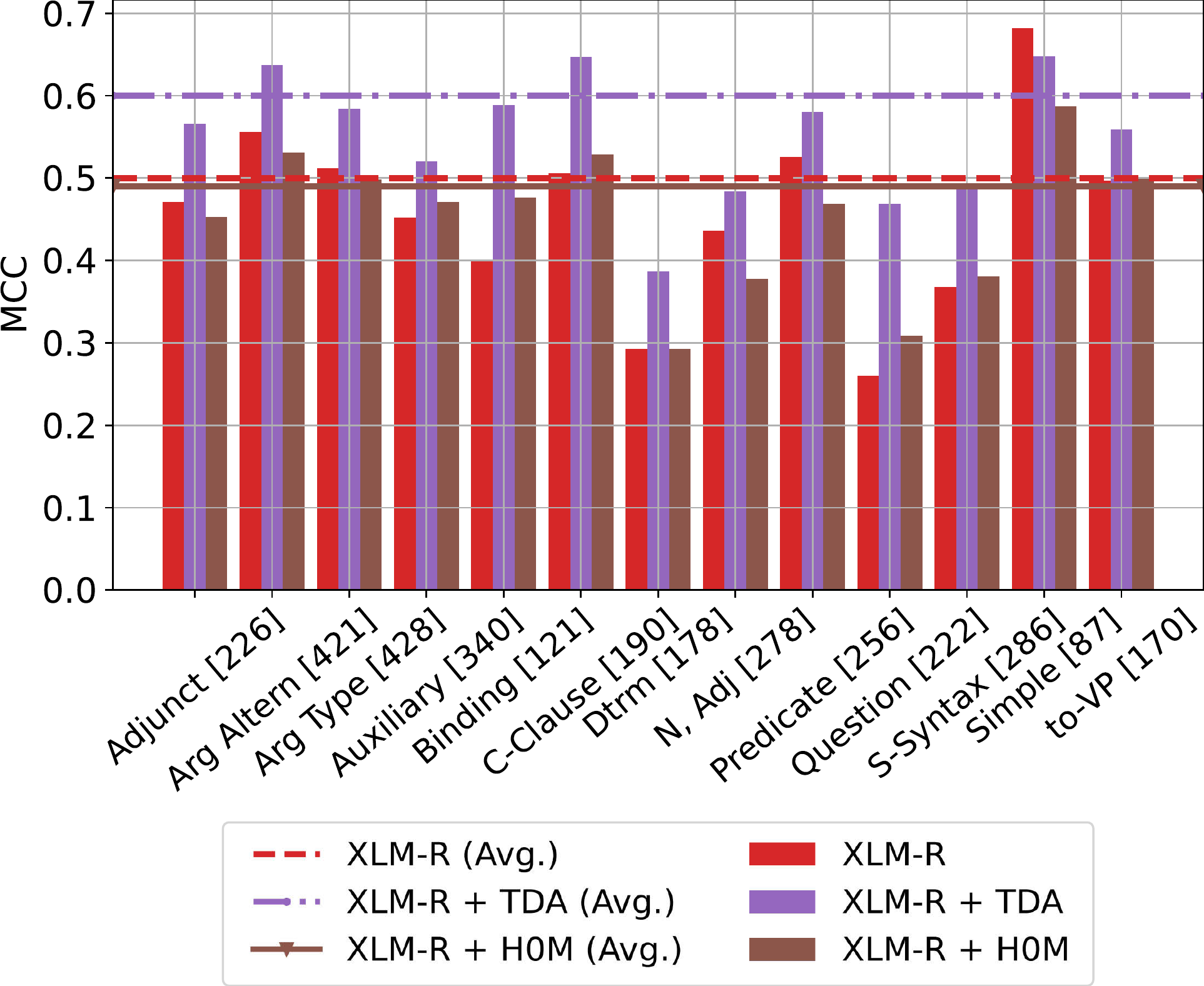}
    \caption{Performance (MCC) of the \emph{fine-tuned} XLM-R by major linguistic feature. Average MCC scores are represented with dashed lines. The number of sentences including the feature is placed in square brackets.}
    \label{fig:cola_xlmr}
\end{figure}

\subsection{Analysis of the Feature Space}
\label{sec:appendix_pca}
We analyze the contribution of the topological features to acceptability classification in the context of linguistic phenomena. We interpret the principal components computed on the \emph{fine-tuned} En-BERT + \textit{TDA} features and identify their importance w.r.t. each head/layer with Shapley values.

\noindent \textbf{Method.} The pipeline assembles the feature  standardization, PCA and training a logistic regression classifier. We conduct a grid search over two pipeline's parameters (i) the number of components $N_{comp}\in[10, 20, \dots, 100]$ (the found optimum: $N_{comp} = 100$) and (ii) regularization parameter of logistic regression $L_{1}\in[0.01, 0.02, \dots, 0.1]$ (the found optimum: $L_{1} = 0.1$). The parameter search is run across $3$ stratified folds, where the \textsc{CoLA} train set is randomly split into train/development sets. The classifier performance is evaluated on the grammatically annotated \textsc{CoLA} development set. We also explore masking principal components, i.e., training the classifier using only the most important components while zeroing the weights of the others.

\autoref{tab:pca_results} shows results for the full pipeline (En-BERT + \textit{TDA} + PCA) and masked pipelines (En-BERT + \textit{TDA} + PC$^1$/PC$^2$). Since the performance is comparable with the En-BERT + \textit{TDA} classifier in \S\ref{sec:acceptability_classification}, we rely on the PCA decomposition for the feature analysis and interpretation.

\begin{table}[t!]
\centering
\small 
\newcommand{\hsp}{\hspace{4pt}}
\begin{tabular}{p{3.1cm}p{1.7cm}p{1.7cm}} 
\toprule
\multirow{2}{*}{\textbf{Model}} &  \textbf{IDD} & \textbf{OODD}  \\[0.4ex]
& \textbf{Acc.} \hsp \textbf{MCC} & \textbf{Acc.} \hsp \textbf{MCC}  \\
\midrule
En-BERT + \textit{TDA}  & 88.6 \hsp 0.725 & 82.1 \hsp 0.565 \\
\midrule
En-BERT + \textit{TDA} + PCA & 84.8 \hsp 0.632  &  81.8 \hsp 0.558 \\
En-BERT + \textit{TDA} + PC$^1$ & 84.1 \hsp 0.609  &  79.1 \hsp 0.482 \\
En-BERT + \textit{TDA} + PC$^2$ & 84.3 \hsp 0.615  &  81.2 \hsp 0.541 \\
\bottomrule
\end{tabular}
\caption{Acceptability classification results with PCA on \textsc{CoLA}. \textbf{IDD}=``in domain dev'' set. \textbf{OODD}=``out of domain dev'' set. Components Set $^1$: \{1\}, Components Set$^2$: \{1,7,9,24,12,0\}.}
\label{tab:pca_results}
\end{table}

\begin{figure}[t!]
    \centering
    \includegraphics[width=\linewidth]{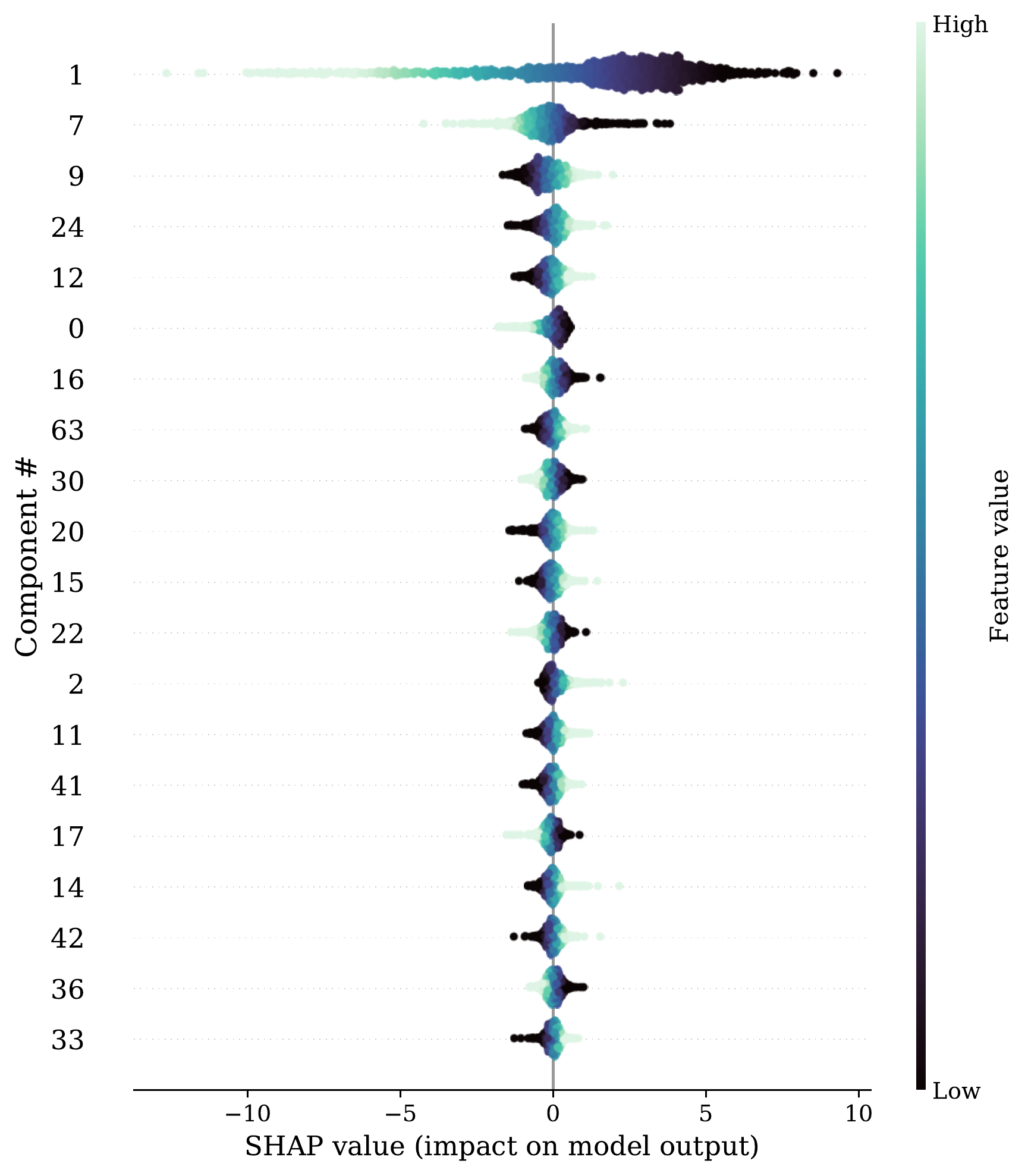}
    \caption{Importance of the PCs for judging sentence acceptability. Shapley values $\phi$ reflect the PCs' impact on the classifier output.}
    \label{fig:pca_importance}
\end{figure}

\begin{figure}[t!]
    \centering
    \includegraphics[width=\linewidth]{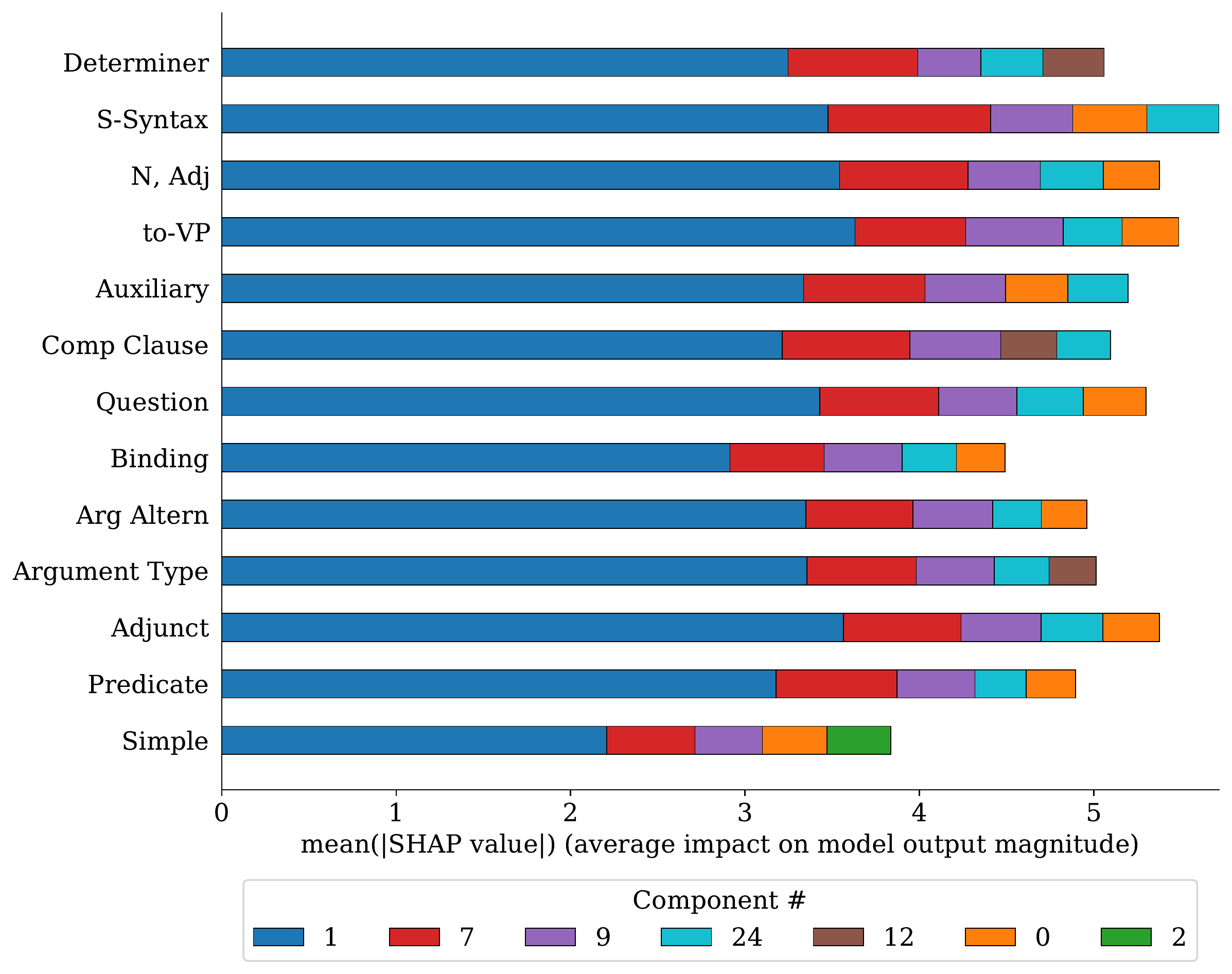}
    \caption{Concatenated mean absolute Shapley values for the important PCs by major linguistic feature.}
    \label{fig:pca_importance_phenomena}
\end{figure}

\vspace{0.1em} \noindent \textbf{Results.}
The following six principal components (PCs) contribute most to acceptability classification according to the mean absolute Shapley values $\phi$ (see~\autoref{fig:pca_importance}). \autoref{fig:pca_importance_phenomena} shows the Shapley values for these PCs by the major linguistic feature.

\noindent{\bf PC1} ($\phi$=$3.179$) has the most impact on the classifiers' performance. {\bf PC1} primarily contains simple topological features (the average vertex degree, the number of edges, and the number of connected components) from the heads at the last layer, which is affected most by the fine-tuning.

\noindent {\bf PC7} ($\phi$=$0.601$) includes same heads as {\bf PC1}, but its features utilize the number of cycles in the attention graph.

\noindent {\bf PC9} ($\phi$=$0.442$) groups all attention patterns except for attention to commas for heads at the lower and middle layers. The component attributes to all phenomena.

\noindent {\bf PC24} ($\phi$=$0.296$) is responsible for topological features at the first and last layers.

\noindent {\bf PC12} ($\phi$=$0.278$) contains the attention to the \texttt{[CLS]} token/next token patterns. The PC is important for classifying sentences including the following features: 
negative polarity and free choice items (\textit{Determiner}), obliques, expletives, prepositional phrases and arguments (\textit{Argument Type}), complement clauses without complementizers
(\textit{Complement Clause}). 

\noindent{\bf PC0} ($\phi$=$0.274$) represents features equal to the number of nodes in the graph, that is, the sentence length in tokens. The PC influences the classifier prediction w.r.t. most of the phenomena, except for \textit{Determiner}, \textit{Complement Clause}, and \textit{Argument Types}.

The following four PCs are less important for acceptability classification ($\phi_{j} < 0.25$) in general but may contribute to some linguistic phenomena.

\noindent{\bf PC16} ($\phi$=$0.243$) comprises topological and distance to pattern features of different heads at the middle layers. The PC contributes to negative polarity and free choice items, non-finite complementizer phrases, and comparative constructions.

\noindent {\bf PC20} ($\phi$=$0.216$) reflects attention-to-comma for various heads at the lower layers. However, this feature helps to classify sentences that fall under the \textit{S-Syntax} and \textit{Question} categories. 

\noindent {\bf PC15} ($\phi$=$0.216$) includes the attention to the first token pattern for the middle-to-higher layer heads (generally 4-to-10). It works for \textit{Passive} and \textit{By-Phrases}.

\noindent{\bf PC2} ($\phi$=$0.203$) reflects the number of graph edges for heads at the first layer, which captures strong pair-wise information about tokens. This head is important for sentences with default syntax  (\textit{Simple}).

\noindent{\bf PC3} and {\bf PC6} ($\phi < 0.05$) represent attention to the dot pattern. The PCs are not important for any of the linguistic phenomena and have large eigenvalues.

\ifx
\begin{table}[]
\centering
\begin{tabular}{@{}lc@{}} 
\toprule
\multirow{1}{*}{\textbf{PC Set}} &  Phenomena  \\
\midrule
[0, 1, 2, 7, 9]  & Simple\\
[0, 1, 7, 20, 24]  &  Psuedo-Aux \\
[0, 1, 7, 9, 14] &  NNCompd \\
[0, 1, 7, 9, 20]  &  Frag/Paren \\
[0, 1, 7, 9, 24]  &  Binding:Other, CP Arg NP/AP, Complex QP, Compx NP, Control, Coord, Copula, Deep Embed,Deverbal, Dislocation, Drop Arg,Ellipsis/Anaphor, Emb Q, Locative, Misc, Modal, NP Adjunct, Neg, PPArg-NP/AP, Particle, Partitive, Quantifier, RC, Rel Adj, Rel NP, S-Adjunct, Temporal, Trans Adj, Trans NP, VP Adjunct, VP arg-NP/AP \\
[1, 7, 16, 20, 22]  &  Comparative \\
[1, 7, 9, 11, 12] &  Result/Depictive \\
[1, 7, 9, 11, 24] &  Island \\
[1, 7, 9, 12, 15]  &  Passive, by-Phrase \\
[1, 7, 9, 12, 16] &  NPI/FCI, Non-finite CP\\
[1, 7, 9, 12, 24] &  Add Arg, Aux, Binding:Refl, Expletive\\
\bottomrule
\end{tabular}
\label{tab:pca_minor_importance}
\caption{Components minor importances}
\end{table}
\fi

\section{Attention Head Selection}
\label{sec:appendix_head_selection}

\subsection{Head Selection Procedure} \label{subsection:head_selection}

We use publicly available scripts\footnote{\href{https://github.com/alexwarstadt/data_generation}{\texttt{github/alexwarstadt/data\_generation}}} to generate up to $100$ minimal pairs per each of $67$ types, ensuring no overlap with the \textsc{BLiMP} pairs. We select the best-performing individual heads and head ensembles by estimating their scoring performance on the generated data and further evaluate them on \textsc{BLiMP}. 

\noindent Algorithms~\ref{best_head}-\ref{cat_head} describe the Top Head and Phenomenon Head selection procedures using a brute force search, while Algorithm~\ref{head_ensemble_selection} presents the process of selecting the Head Ensembles via beam search.

\begin{algorithm}[th!]
    \scriptsize
	\caption{Top Head Selection} 
    \label{best_head}

	\begin{algorithmic}[2]
	\pseudoINPUT{Set $Q_1$:  contains all possible pairs $ (h, r) $, where $h$ -- attention head and $r \in \{1, 2\}$ -- scoring rule}
	\Require {$acc(.):$ accuracy evaluation function of the head with the selected rule on pairs for all phenomenas}
	\pseudoOUTPUT {Pair $(H_B, R_B)$}
	\Procedure{Selecting Top Head}{$Q_1$}
	\State $BestAcc \leftarrow 0$
	\State $(H_B, R_B) \leftarrow (-1, -1)$
	\For{$(h, r) \in Q_1$}
	    \If{$acc((h, r)) > BestAcc$}
	    \State $BestAcc \leftarrow acc_C((h, r))$
	    \State $(H_B, R_B) \leftarrow (h, r)$
	    \EndIf
	\EndFor
	\State{\textbf{return: }  $(H_B, R_B)$}
	\EndProcedure
	\end{algorithmic}
\end{algorithm}

\begin{algorithm}[h!]
    \scriptsize
	\caption{Phenomenon Head Selection} 
    \label{cat_head}

	\begin{algorithmic}[2]
	\pseudoINPUT{Set $Q_1$:  contains all possible pairs $ (h, r) $, where $h$ -- attention head and $r \in \{1, 2\}$ -- scoring rule}
	\Require{$C:$ linguistic category}
	\Require {$acc_C(.):$  accuracy evaluation function of the head with the selected scoring rule on the $C$ pairs}
	\pseudoOUTPUT {Pair $(H_C, R_C)$}
	\Procedure{Selecting Phenomenon Head}{$Q_1$}
	\State $BestAcc \leftarrow 0$
	\State $(H_C, R_C) \leftarrow (-1, -1)$
	\For{$(h, r) \in Q_1$}
	    \If{$acc_C((h, r)) > BestAcc$}
	    \State $BestAcc \leftarrow acc_C((h, r))$
	    \State $(H_C, R_C) \leftarrow (h, r)$
	    \EndIf
	\EndFor
	\State{\textbf{return: }  $(H_C, R_C)$}
	\EndProcedure
	\end{algorithmic}
\end{algorithm}

\begin{algorithm}[t!]
    \scriptsize
	\caption{Head Ensemble Selection} 
    \label{head_ensemble_selection}
	\begin{algorithmic}[2]
	\pseudoINPUT{Set $Q_1$:  contains all possible pairs $ (h, r) $, where $h$ -- attention head and $r \in \{1, 2\}$ -- scoring rule}
	\Require {$acc(.):$  accuracy evaluation function of the ensembles with selected scoring rules using majority voting}
	\pseudoOUTPUT {Ensemble $B$: set of pairs $(H, R)$}
	\Procedure{Selecting Head Ensemble}{$Q_1$}
	\State $Q \leftarrow \lbrace \lbrace (h, r) \rbrace \mid (h, r) \in Q_1\rbrace $
	\Do 
        \State $Q^\prime \leftarrow \emptyset$
        \For{$q \in Q$}
            \For{$(h_1, r_1) \neq (h_2, r_2) \in Q_1 \setminus q$}
            \State $q^\prime \leftarrow q \cup \{(h_1, r_1), (h_2, r_2)\}$
            \If {$acc(q^\prime) > acc(q)$}
                \State $Q^\prime \leftarrow Q^\prime \cup \lbrace q^\prime\rbrace $
            \EndIf 
            \EndFor
        \EndFor
	\If {$\rvert Q\lvert ~ \ge 40$}
	    \State $Q \leftarrow $ top-$40$ pairs $q$, scored by $acc(.)$,  $q \in Q^\prime$
	\Else 
	    \If {$\rvert Q^\prime \lvert ~ > 0$}
	        \State $Q \leftarrow Q'$
	   \EndIf
	\EndIf
	\doWhile{$Q^\prime \ne \emptyset$}
	
	\State{\textbf{return: }  Ensemble from $Q$}
	\EndProcedure
	\end{algorithmic}
\end{algorithm}

\begin{figure*}[!h]
    \centering
    \includegraphics[width=.9\linewidth]{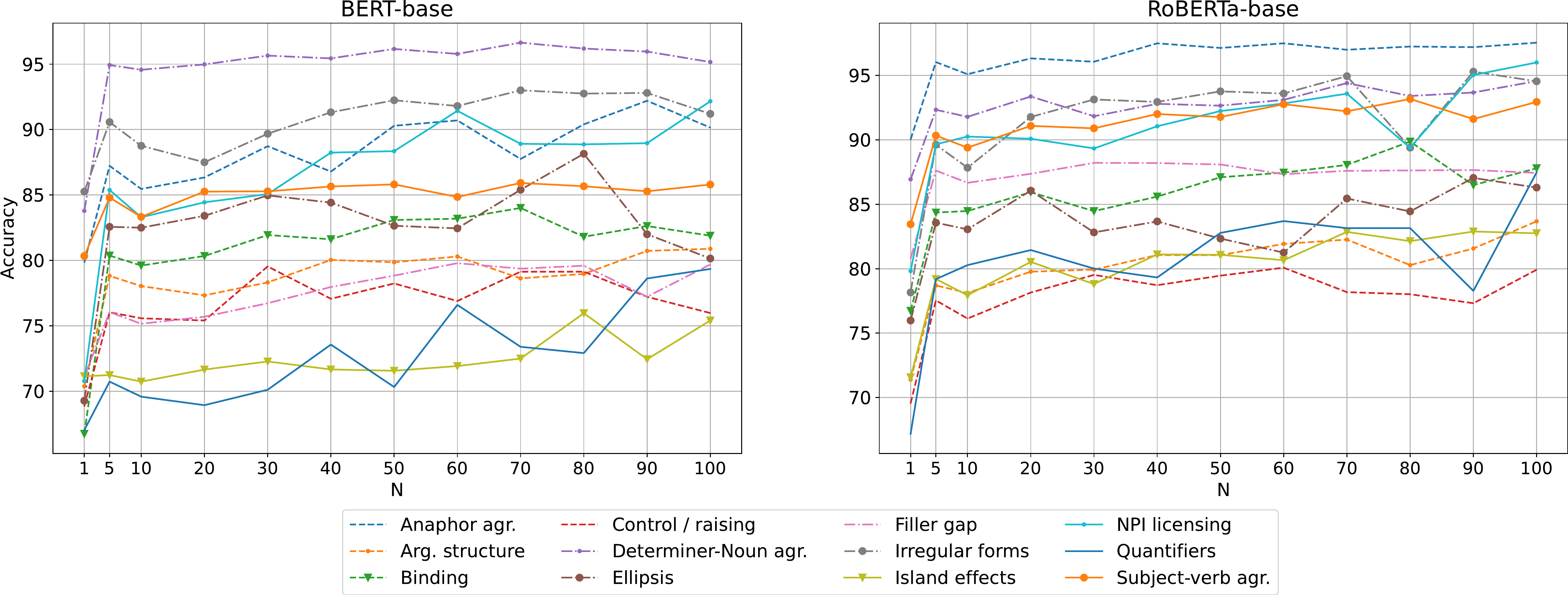}
    \caption{The effect of a given amount of examples on the \textsc{BLiMP} performance of selected Head Ensembles by major category. Method=RTD scoring. N=number of extra examples per phenomenon.}
    \label{fig:models_blimp_n}
\end{figure*}

\subsection{Effect of Auxiliary Data} \label{subsection:auxdata}
We analyze the effect of the amount of auxiliary generated data on the RTD scoring performance. We explore $N \in [1, 5, 10, ..., 100]$ sentence pairs per language phenomenon used for selecting Head Ensembles as described in Appendix~\ref{subsection:head_selection}. The experiments are run ten times, where each run includes generation of auxiliary minimal pairs, the corresponding head selection procedure, and evaluation on \textsc{BLiMP} for each and all language phenomena. The accuracy performance is averaged over all experiment runs.

\vspace{0.3em} \noindent \textbf{Results.} \autoref{fig:models_blimp_n} presents the results for the BERT-base and RoBERTa-base models. We observe that some phenomena receive prominent performance using only one auxiliary example (e.g., BERT-base: \emph{Irregular forms}: $85\%$; \emph{Determiner-Noun agr}: $84\%$; \emph{Anaphor agr}: $80\%$; \emph{Subject-verb agr}: $80\%$; RoBERTa-base: \emph{Anaphor agr}: $90\%$; \emph{Determiner-Noun agr}: $87\%$; \emph{Subject-verb agr}: $83\%$). Both models receive similar maximum scores for some phenomena with a given different amount of examples (\emph{Control / raising}: $79\%$/$80\%$, \textbf{N}=$30/60$; \emph{Ellipsis}: $88\%$/$87\%$, \textbf{N}=$80/90$; \emph{Filler gap}: $80\%$/$88\%$, \textbf{N}=$60/30$). By contrast, there is a significant difference between the maximum and minimum scores on certain phenomena, e.g., \emph{Quantifiers} ($10\%$/$9\%$), \emph{NPI licensing} ($8\%$/$6\%$), and \emph{Ellipsis} ($8\%$/$6\%$).

\section{The $H_0S$ Feature Distributions}
\label{appendix:h0s}
\begin{figure}[h!]
  \centering
  \includegraphics[width=.93\linewidth]{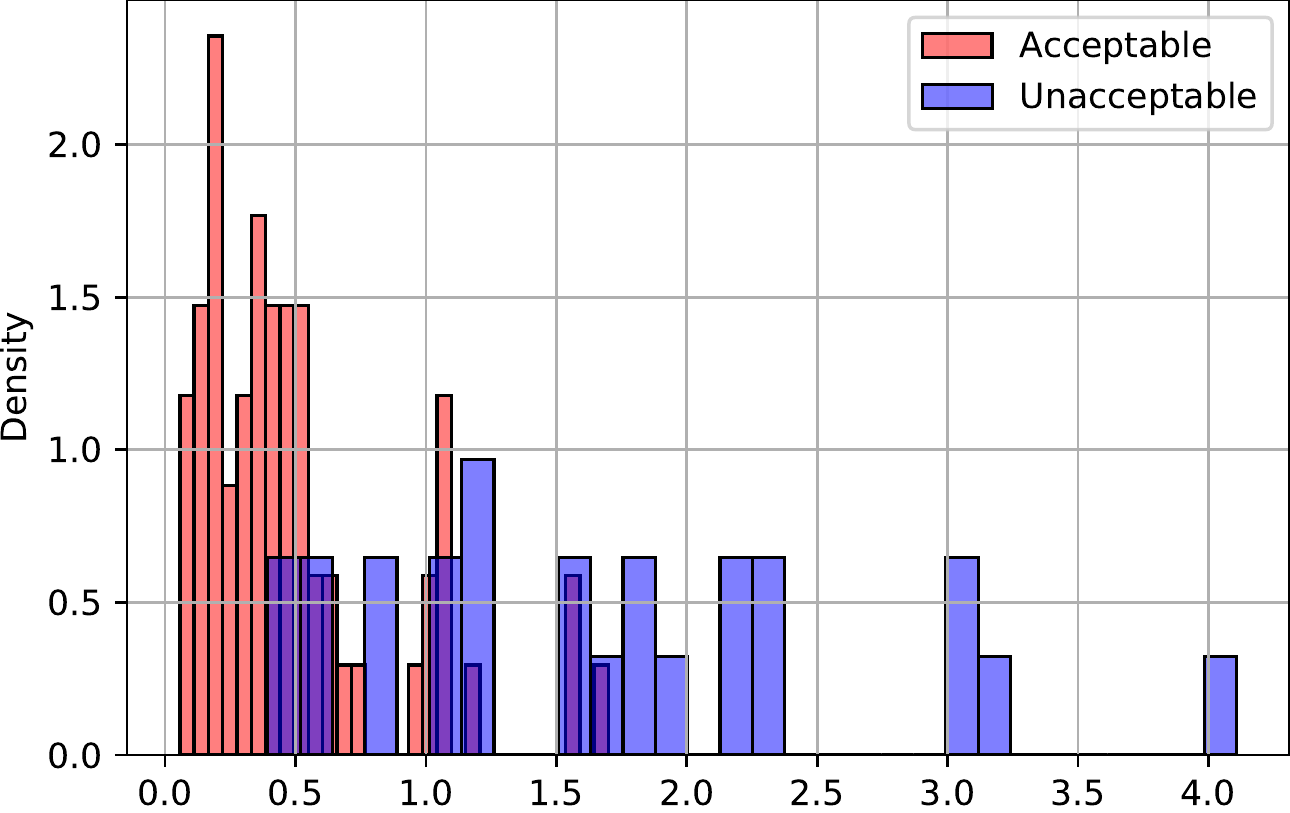}
  \caption{The distribution shift of the $H_0S$ feature between the acceptable and unacceptable sentences (\textit{Simple}); [L: $10$; H: $3$].}
  \label{fig:h0s_simple}
\end{figure}

\begin{figure}[h!]
  \centering
  \includegraphics[width=.93\linewidth]{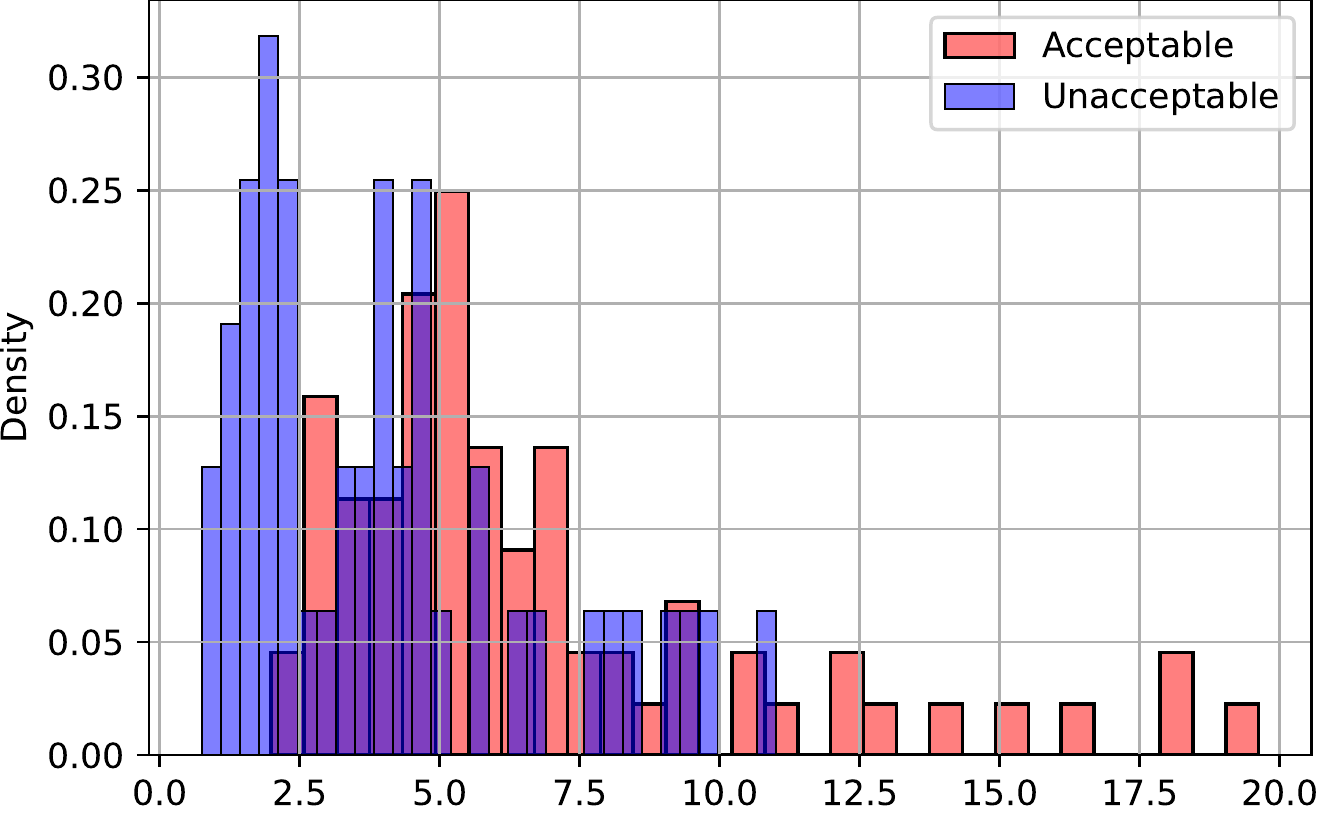}
  \caption{The distribution shift of the $H_0S$ feature between the acceptable and unacceptable sentences (\textit{Binding}); [L: $11$; H: $4$].}
  \label{fig:h0s_binding}
\end{figure}

\autoref{fig:h0s_simple} and \autoref{fig:h0s_binding} illustrate examples of the $H_0S$ feature distribution shifts between the acceptable and unacceptable sentences from the entire \textsc{CoLA} development set.


\section{Toy Examples of Calculating Features}
\label{appendix:example}
\begin{figure*}
\captionsetup[subfigure]{justification=centering}
    \centering
    \begin{subfigure}[t]{0.24\textwidth}
        \includegraphics[width=\textwidth]{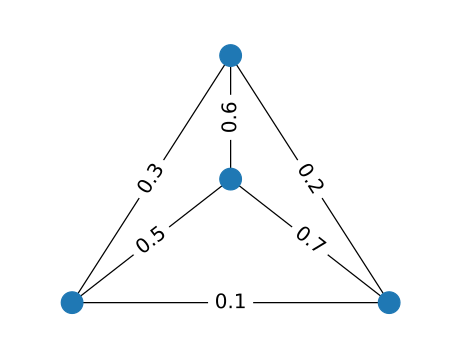}
        \caption{ $G_{toy}$}
        \label{fig:graph_example}
    \end{subfigure}
    ~ %
    \begin{subfigure}[t]{0.24\textwidth}
        \includegraphics[width=\textwidth]{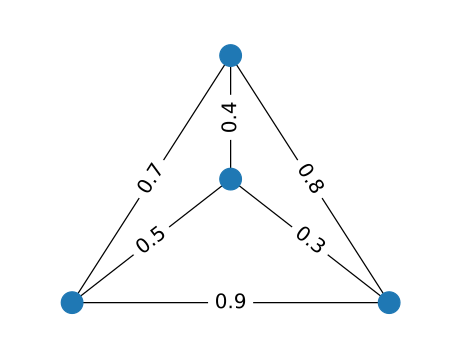}

        \caption{ $G_{toy}'$}
        \label{fig:graph_ripser_example}
    \end{subfigure}
    \begin{subfigure}[t]{0.24\textwidth}
        \includegraphics[width=\textwidth]{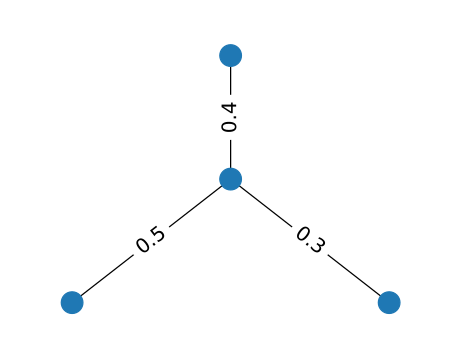}
        \caption{minimum spanning  tree \\ of~$G_{toy}'$}
        \label{fig:graph_mst_example}
    \end{subfigure}
    \begin{subfigure}[t]{0.24\textwidth}
        \includegraphics[width=\textwidth]{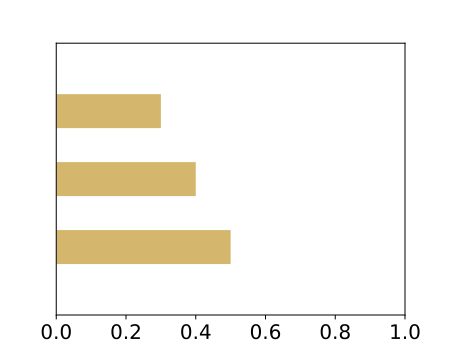}
        \caption{$H_0$-barcode of $G'_{toy}$}
        \label{fig:barcode_example}
    \end{subfigure}
    \caption{An example of a weighted graph and corresponding $H_0$-barcode, calculated with the Ripser++ library.}\label{fig:simple_example}
\end{figure*}

Let us demonstrate a calculation of essential barcode features of a toy graph $G_{toy}$ ( \autoref{fig:graph_example}). First, we calculate the $H_0$-barcode of this graph. To do it, we build the graph $G'_{toy}$ by replacing each edge weight $w$ with $1 - w$, as in \autoref{fig:graph_ripser_example}. Next we calculate the minimum spanning tree of this new graph (\autoref{fig:graph_mst_example}). We end up with the $H_0$-barcode with the lengths of bars, equal to the weights of the minimum spanning tree  (\autoref{fig:barcode_example}). We can derive $H_0S(G_{toy}) = 0.3 + 0.4 + 0.5 = 1.2$ and $H_0M(G_{toy}) = H_0S(G_{toy}) / 3 = 0.4$ from this barcode diagram.

Note that the directions of the bars (\autoref{fig:graph_examples}) were reversed comparing to the actual ripser++ output (\autoref{fig:barcode_example}). The reversed representation is more intuitive: edges with lower weights are filtered out earlier than edges with the higher weights.

Next we compute the Betti numbers for the same graph $G_{toy}$ given three thresholds: $\tau_1 = 0$, $\tau_2 = 0.4$ and $\tau_3 = 1$. 

At $\tau = 0$ we do not drop any edges. We have the full graph with one connected component (\autoref{fig:graph_example}).  $\beta_0$ is defined as the number of connected components. Hence $\beta_0$ equals to  $1$. Next we calculate $\beta_1$, using the shortcut formula for graphs: $\beta_1 = |E| + |C| - |V|$. In our case,  $|E| = 6$ is the number of edges, $|C| = 1$ is the number of connected components and $|V| = 4$ is the number of vertices. Finally, we get $\beta_1 = 3$. Note that $\beta_1$ corresponds to three simple undirected loops in the graph. %
There is also an alternative method to represent the graph and to  calculate the first Betti number. This method does not account for  ``trivial'' loops, which are defined by the  triangles borders. It was used in the example above, see \autoref{fig:graph_examples}. %

At $\tau = 0.4$, we drop all edges with weights lower than $0.4$. We get the same structure as the minimum spanning tree of the graph $G_{toy}'$ (\autoref{fig:graph_mst_example}), but without weights inversion. For this graph, $\beta_0 = 1$: there is a single connected component,  $\beta_1 = 3 + 1 - 4 = 0$. It corresponds  to the number of simple loops, which equals to $0$.

At $\tau = 1$, we drop all edges as  all edges have weights below than $1$. The resulting graph consists only of vertices without edges. For this case, we have four connected components, so $\beta_0 = 4$, and $\beta_1 = 0 + 4 - 4 = 0$.

\end{document}